\documentclass[11pt]{article}
\usepackage[utf8]{inputenc}
\usepackage{geometry}
\usepackage{microtype}
\usepackage[T1]{fontenc}
\usepackage{graphicx}
\usepackage{calc,ifthen,setspace,amsmath,amscd,amssymb}
\usepackage{multirow}
\usepackage{algorithm}
\usepackage{algorithmic}
\usepackage{subcaption}
\usepackage{tikz}
\usepackage{url}            
\usepackage{float}
\usepackage{bm}
\usepackage{blindtext}
\usepackage[square,super,comma,authoryear]{natbib}
\usepackage{upgreek}
\usepackage{hyperref}
\usepackage[noabbrev,capitalize,nameinlink]{cleveref}
\usepackage[normalem]{ulem}

\newcommand{\E}{\mathbb{E}}

\newcommand{\Norm}{\text{N}} 
\newcommand{\Unif}{\text{Unif}} 

\newcommand{\bW}{\bm{W}}
\newcommand{\bw}{\bm{w}}
\newcommand{\by}{\bm{y}}
\newcommand{\bx}{\bm{x}}

\newcommand{\bz}{\bm{z}}

\newcommand{\bb}{\bm{b}}

\newcommand{\btheta}{\bm{\theta}}

\DeclareMathOperator{\Var}{Var}

\newcommand{\Data}{\bm{\mathcal{D}}}

\newcommand{\bmu}{\bm{\mu}}

\newcommand{\loo}{ \widehat{\text{elpd}}_{\text{loo}}}
\newcommand{\RMSE}{\text{RMSE}}
\newcommand{\dif}{\text{d}}

\begin{document}

\title{Understanding the Trade-offs in Accuracy and Uncertainty Quantification: Architecture and Inference Choices in Bayesian Neural Networks} 

\author{Alisa Sheinkman\thanks{School of Mathematics and Maxwell Institute for Mathematical Sciences, University of Edinburgh, \href{mailto:a.sheinkman@sms.ed.ac.uk }{a.sheinkman@sms.ed.ac.uk }, \href{mailto:sara.wade@ed.ac.uk}{sara.wade@ed.ac.uk}} \, and \, Sara Wade$^{*}$}


\maketitle              

\begin{abstract}
As modern neural networks get more complex, specifying a model with high predictive performance and sound uncertainty quantification becomes a more challenging task. Despite some promising theoretical results on the true posterior predictive distribution of Bayesian neural networks, the properties of even the most commonly used posterior approximations are often questioned. Computational burdens and intractable posteriors expose miscalibrated Bayesian neural networks to poor accuracy and unreliable uncertainty estimates. Approximate Bayesian inference aims to replace unknown and intractable posterior distributions with some simpler but feasible distributions. The dimensions of modern deep models, coupled with the lack of identifiability, make Markov chain Monte Carlo (MCMC) tremendously expensive and unable to fully explore the multimodal posterior. On the other hand, variational inference benefits from improved computational complexity but lacks the asymptotical guarantees of sampling-based inference and tends to concentrate around a single mode.
The performance of both approaches heavily depends on architectural choices; this paper aims to shed some light on this, by 
considering the computational costs, accuracy and uncertainty quantification in different scenarios including large width and out-of-sample data. To improve posterior exploration, different model averaging and ensembling techniques are studied, along with their benefits on predictive performance. In our experiments, variational inference overall provided better uncertainty quantification than MCMC; further, stacking and ensembles of variational approximations provided comparable accuracy to MCMC  at a much-reduced cost.

\end{abstract}

\section{Introduction}  \label{sec:experimentsintro}
 Despite the tremendous success of deep learning in areas such as natural language processing \citep{touvron2023llamaopenefficientfoundation} and computer vision \citep{sutskever2017,dosovitskiy2021imageworth16x16words}, often there is no clear understanding of why a particular model performs well \citep{zhang2021understanding,szegedy2013intriguing}. Even though the universal approximation theorem guarantees that a wide enough feed-forward neural network with a single hidden layer can express any smooth function \citep{hornik1989}, in practice, constructing a model which is not only expressive but generalizes well is challenging. In contrast, the so-called no free lunch theorem \citep{Wolpert} dictates that there is no panacea to solve every problem, and one should be careful when designing a model appropriate to the task. 
Many modern machine learning models are over-parametrized and prone to overfitting, especially given the limited size of the dataset. Complex problems demand exploring bigger model spaces, and there is a danger of choosing an excessively over-parametrized model, which is going to overfit and have a high variance. Additionally, conventional deep models do not offer human-understandable explanations and lack interpretability \citep{lipton2018mythos}. By default, classical neural networks do not address the uncertainty associated with their parameters and whilst there exist proposals enabling neural networks (NNs) to provide some uncertainty estimates, they are often miscalibrated \citep{guo17}. As a result, these models are typically overconfident, provide a low level of uncertainty even when data variations occur \citep{Ovadia2019canyoutrust}, and are easily fooled and are susceptible to adversarial attacks \citep{szegedy2013intriguing,nguyen2015deep}. At the same time, reliable uncertainty quantification (UQ) is crucial for any decision-making process, and it is not enough to obtain a point estimate of the prediction.  

The key distinguishing property of the Bayesian framework is that it incorporates domain expertise and deals with uncertainty quantification in a principled way: by marginalizing with respect to the posterior distribution of parameters. As a result, Bayesian models are more resistant to distribution shifts and can improve the accuracy and calibration of classical deep models \citep{wilson2020bayesian}. Nevertheless, the reliability of uncertainty estimates and the gap between within-the-sample and out-of-sample performance still require improvement \citep{foong2020expressiveness}. The posterior distributions arising in Bayesian neural networks (BNNs) are analytically unavailable and highly multimodal, and the core challenge lies in estimating the posterior \citep{papamarkou2022a}. One should not only find a model that matches the task but, as importantly, achieve the alignment between the model and the applied inference algorithm \citep{gelman2020bayesianworkflow}; and the most theoretically grounded sampling methods and approximation techniques are limited by the computing budget, size of the dataset, and sheer number of parameters. 
We list several characteristics of classical and Bayesian neural networks in the \cref{tab:chalofDLandBDL}. 
\begin{table}[t!]
\centering
\caption{Some of the challenges and properties of classical and Bayesian neural networks.}
\label{tab:chalofDLandBDL}
\begin{tabular}{l|l|l}
Property & Classical NN  & Bayesian NN \\ \hline \hline
Interpretability & poor  & improved $\checkmark$\\ \hline
Robustness to OOD  & poor  & improved $\checkmark$ \\ \hline
Adversarial attacks & sensitive & less sensitive $\checkmark$ \\ \hline
Overconfidence & typical & less typical $\checkmark$ \\ \hline
Training outcome  & point estimate $\cdot$ & posterior distribution $\mathbb{P}$ \\ \hline
Incorporate prior  & no & yes     \\ \hline
Require initialization & yes & yes \\ \hline
\end{tabular}
\end{table}

\textbf{Outline.} In this work, we consider some of the challenges and nuances of Bayesian neural networks and evaluate the performance with different architectures and for different posterior inference algorithm choices.  Specifically, we study the sensitivity of BNNs to the choice of width in \cref{sec:limitwidth}, depth in \cref{sec:limitdepth}, and investigate the performance of BNNs under the distribution shift in \cref{sec:ood}. Across all the experiments in \cref{sec:experimentsection}, we observe that for different inference algorithms, one model can provide strikingly diverse performances.  The challenge of comparative model assessment is addressed in \cref{sec:elpd}, where we introduce the estimated pointwise loglikelihood as a measure of model utility. While given some set of models, the Bayesian approach has the potential to deal with the model choice by comparing posterior model probabilities, such comparison tends to favour one candidate disproportionally strongly \citep{oelrich2020bayesian}. Thus, the classical Bayesian model averaging (BMA) based on model probabilities \citep{hoeting1999bayesian} is only optimal if the true model is among the comparison set. In response to the limitations of BMA, in \cref{sec:stacking,sec:stackingexample,sec:NASA} we consider ensembling, stacking and pseudo-BMA\citep{Yao_2018}.

\section{Empirical Study of Limiting Scenarios} \label{sec:experimentsection}
\subsection{Architecture Components} 
Whilst the dimensions of the input and the output are determined by the dimensionality of the data set, the dimension of the weight space plays an essential part in specifying neural networks and can be tuned to improve prediction performance. In the case of feed-forward neural networks, this amounts to finding optimal depth and width. While the universal approximation theorem advocates for single-layer neural networks \citep{hornik1989}, variants of the universal approximation theorem exist for deeper networks \citep{Lu2017,hanin2019universal}. Further, deep neural networks gained popularity due to their expressiveness and tremendous success in real-world applications, allowed by the increase in available computing power \citep{chatziafratis2020better}. At the same time, the more parameters one has, the more nuanced the choice of the model becomes. No matter what the prediction task is, overly complex models suffer from the curse of dimensionality which causes not only poor performance but also computational problems. 

On a slightly different line, we recall the seminal result first obtained for neural networks with one hidden layer \citep{neal1995} and then extended to arbitrary depth \citep{matthews} which states that under general conditions, as the width of a BNN tends to infinity, the distribution of the network's output induced by the prior converges to a Gaussian process (GP) with a neural network kernel, also known as the neural network GP (NNGP); there is a similar correspondence relating  GPs and distributions induced by the posterior \citep{hron2020exactposteriordistributionswide}. 
When defining BNNs, choosing a prior and understanding how properties and prior beliefs on the weight space translate to the functions is a major challenge. 
Note that generally, we require priors which are: (1) interpretable, e.g. we want to be able to specify the hyperparameters of the prior based on the task at hand; (2) have large support, i.e. prior should not concentrate around a small subset of the parameter space; (3) lead to feasible inference and favour reasonable approximations of the posterior and predictive distributions.  

Finally, to specify any neural network, one needs to choose the activation function, which (apart from being nonlinear) is required to be differentiable. In our experiments, we consider the widely-used rectified linear unit function (ReLU) defined as $\max(0,x)$, which switches the negative inputs off and leaves the positive ones unchanged, as well as the sigmoid activation function defined as $\sigma(x) = \exp(x)/(\exp(x) + 1)$. 

\subsection{Setup of the Experiments} \label{sec:settingsexperiment}
\begin{figure}[t!]
    \centering
      \subcaptionbox{\centering Example of the DAG of the neural network when $L=2$. \label{dagofbnnforexperiment} }[.6\textwidth]{ 
\includegraphics[width = 0.9\linewidth]{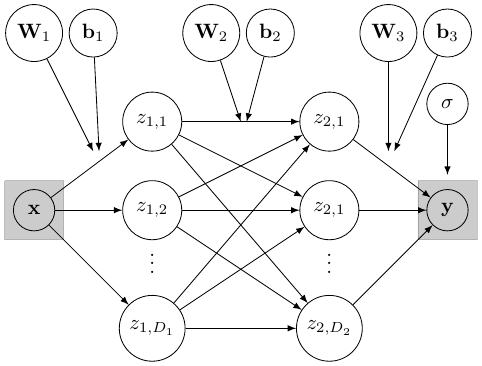}
    }
          \subcaptionbox{\centering Examples of priors of $\bW_l$ for $l = 2, \ldots,L+1$ and $D_l = 20$. \label{weights} }[.39\textwidth]{ 
\includegraphics[width = \linewidth]{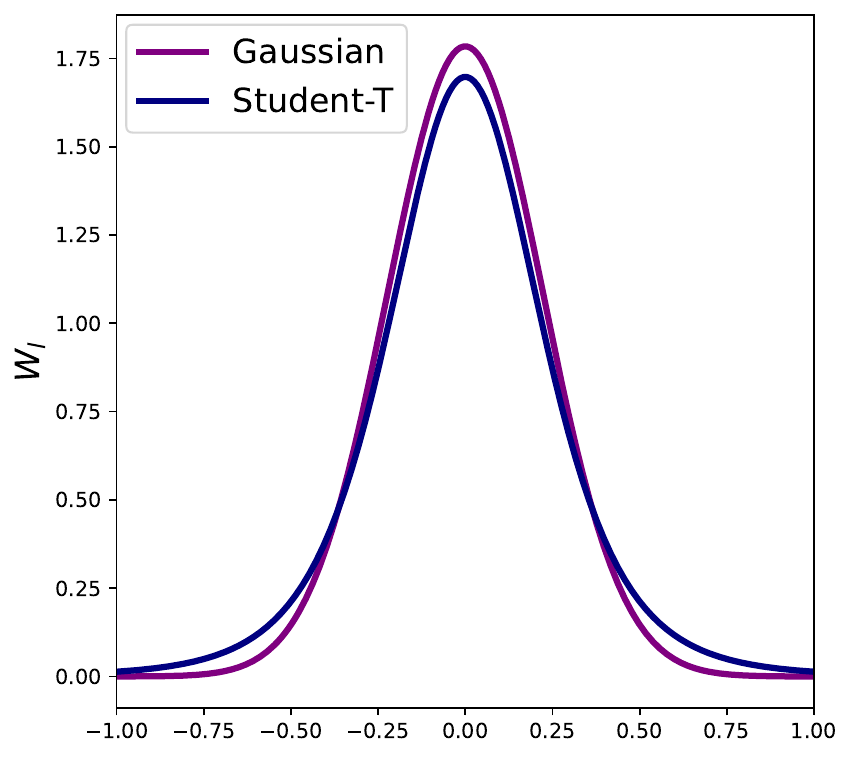}
    }
    \caption{Examples of the directed acyclic graph (DAG) of the neural network and of the priors used in the experiments.}
    \label{fig:priorsdagofbnnforexperiment}
\end{figure}

In the experiments, we consider the following BNN, illustrated by the \cref{dagofbnnforexperiment}
\begin{align}
& \by \sim \Norm \left (\bb_{L+1} + \bW_{L+1}\bz_{L}, \bm{\sigma}\right), \quad \bm{\sigma} \sim |\Norm(0, 0.001)| \label{eq:model_for_experiments} \\
&\bz_l  = g\left ( \bb_{l} + \bW_{l}\bz_{l-1}\right) \text{ for } l = 1, \ldots,L \nonumber, 
\end{align}
 where we consider two different choices of activations $g$, namely, the ReLU and the sigmoid; $|\Norm(,)|$ denotes a half-normal distribution; and $\bz_{0} = \bx$. We consider two possible choices of priors on the weights (illustrated by \cref{weights}): (i) Gaussian priors as the most conventional choice \citep{arbel2023primer,fortuin2022priors}; (ii) Student-t priors, motivated by the observation that empirical weight distributions of SGD-trained networks are heavy-tailed \citep{fortuin2021priors,Gurbuzbalaban2021}.  We finish specifying the model by placing Gaussian priors on the biases, that is:
\begin{align*}
&\bW_1\sim F\left(0, \frac{1}{4L}\right), \quad \bW_l\sim F\left(0, \frac{4}{D_{l-1}}\right) \quad \text{for} \quad l = 2, \ldots,L+1,\\ 
& \bb_l \sim \Norm\left(0,\frac{1}{4L}\right) \quad \text{for} \quad l = 1, \ldots,L+1, 
\end{align*}
where the notation $F(\mu, \sigma^2)$ represents a distribution with mean $\mu$ and scale $\sigma$, and specifically, here is chosen as either Gaussian or Student-t with 5 degrees of freedom.  
To avoid divergence in wider networks and mitigate the damage caused by the nonlinear deformation \citep{he2015delvingdeeprectifierssurpassing}, the weights' variance is scaled by the inverse of the preceding layer's width. 

The BNN defined by \cref{eq:model_for_experiments} and trained with automatic differentiation variational inference (ADVI) \citep{kucukelbir2016advi}, which assumes a mean-field (diagonal) Gaussian variational family, is referred to as mfVIR or mfVIS, depending on the choice of the activation: ReLU or sigmoid, respectively. The model trained with the Hamiltonian Monte Carlo (HMC), using the No U-Turn Sampler (NUTS) \citep{hoffman14gelman} is denoted as HMCR or HMCS.
For simplicity, we often refer to one-layer neural networks of particular width $D$ as to mfVIRD, mfVISD, HMCRD or HMCSD (e.g. one-layer BNN with 20 hidden units and ReLU activation trained with mean-field VI is called mfVIR20). All experiments are implemented with Numpyro \citep{numpyrophan2019composable}, ArviZ \citep{arviz_2019}, JAX \citep{jax2018github} and Flax \citep{flax2020github}. We record the run time of the approximate inference (TT), the root mean squared error $\RMSE$ and empirical coverage for the function and observations (EC). Note that we compute empirical coverage as a fraction of observations contained within the $95\%$ confidence interval (CI), this means that in the ideal settings the computed EC should equal to $0.95$. If $\text{EC}>0.95 $ then the confidence intervals are too wide; a worse scenario occurs when $\text{EC} <0.95 $ as it means that the CIs are too narrow and the model is overconfident in predictions. Details on the computed metrics and the corresponding formulas are discussed in \cref{appendix:metrics}, where we provide further information on the initialization and parameters for the inference algorithms\footnote{The code is available on \href{https://github.com/sheinkmana/ArchitectureofBNNs}{GitHub}.}.

 The absence of the test log-likelihood among the recorded metrics is motivated by the observation that the higher test log-likelihood does not necessarily correspond to a more accurate posterior approximation nor to lower predictive error (such as $\RMSE$)\citep{deshpande2024usingtestloglikelihoodcorrectly}. 

\subsection{Increasing the Width of the Network}
\label{sec:limitwidth}
\begin{figure}[t!]
    \centering
    \subcaptionbox{\centering Metrics of methods as the number of hidden units increases. \label{onelayermetrics} }[.4\textwidth]{ 
\includegraphics[width = \linewidth]{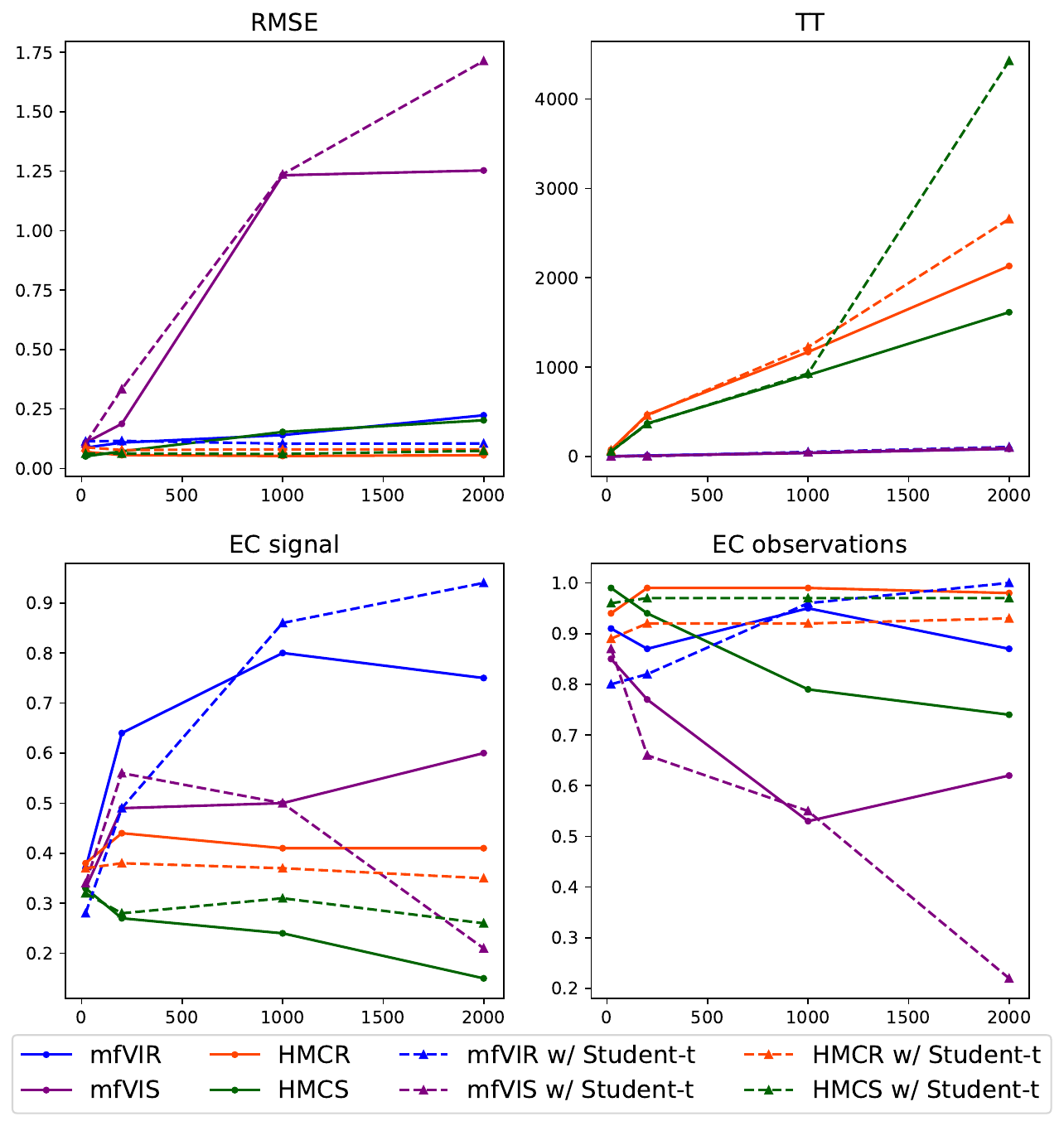}
    }
    \subcaptionbox{\centering Predictions and uncertainty estimates for each method with $D_1=2000$ and Gaussian priors.\label{2000plots} }[.59\textwidth]{ 
    \includegraphics[width = 0.49\linewidth]{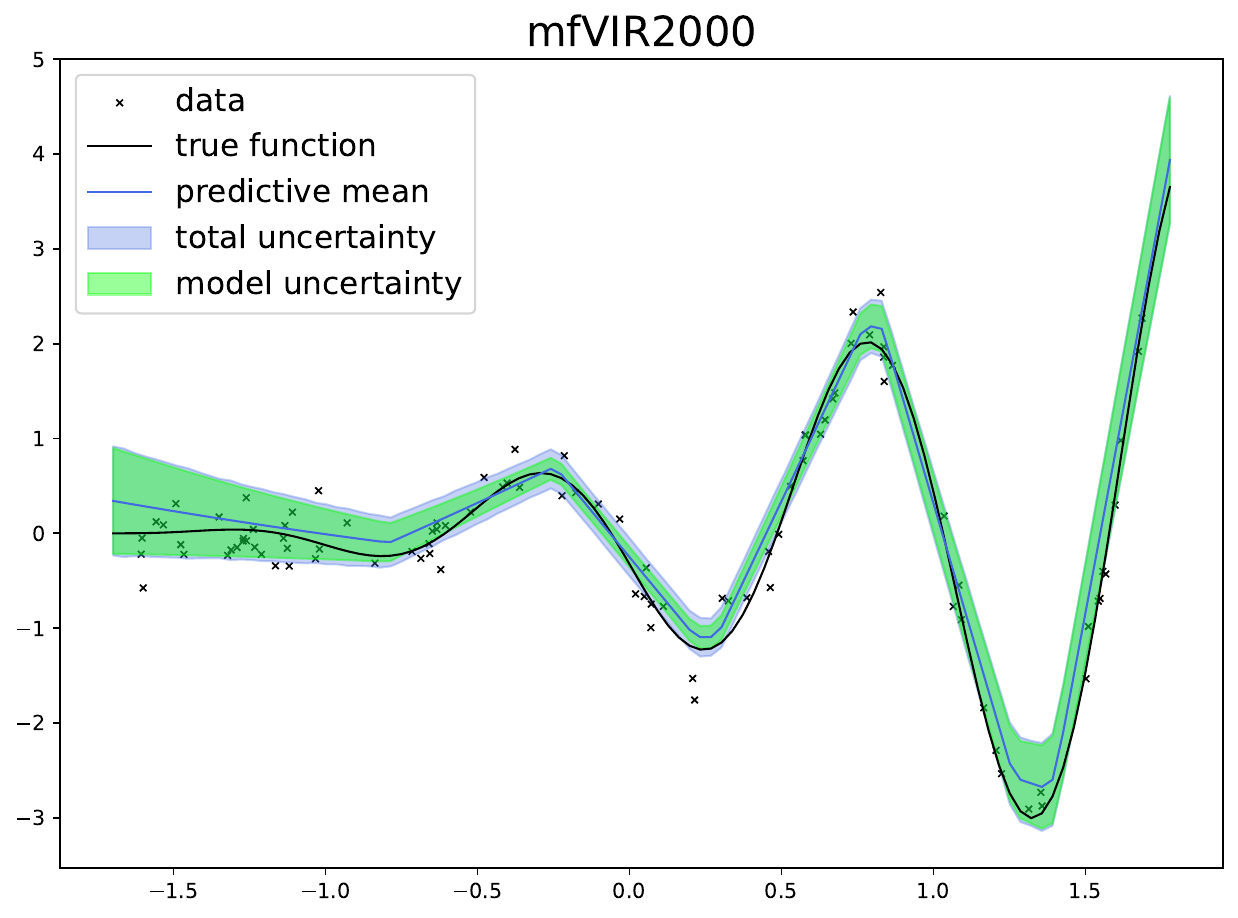}
    \includegraphics[width = 0.49\linewidth]{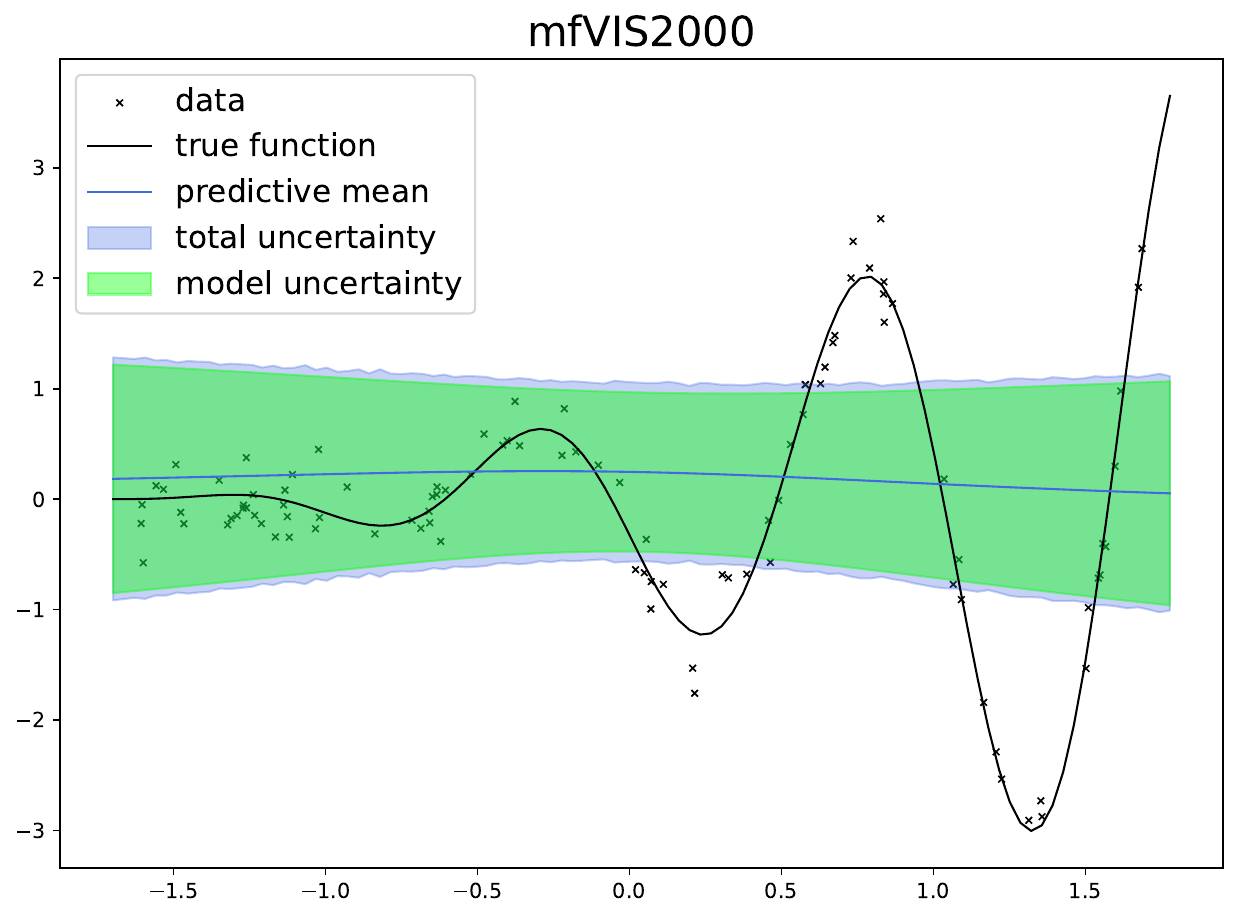} \\
        \includegraphics[width = 0.49\linewidth]{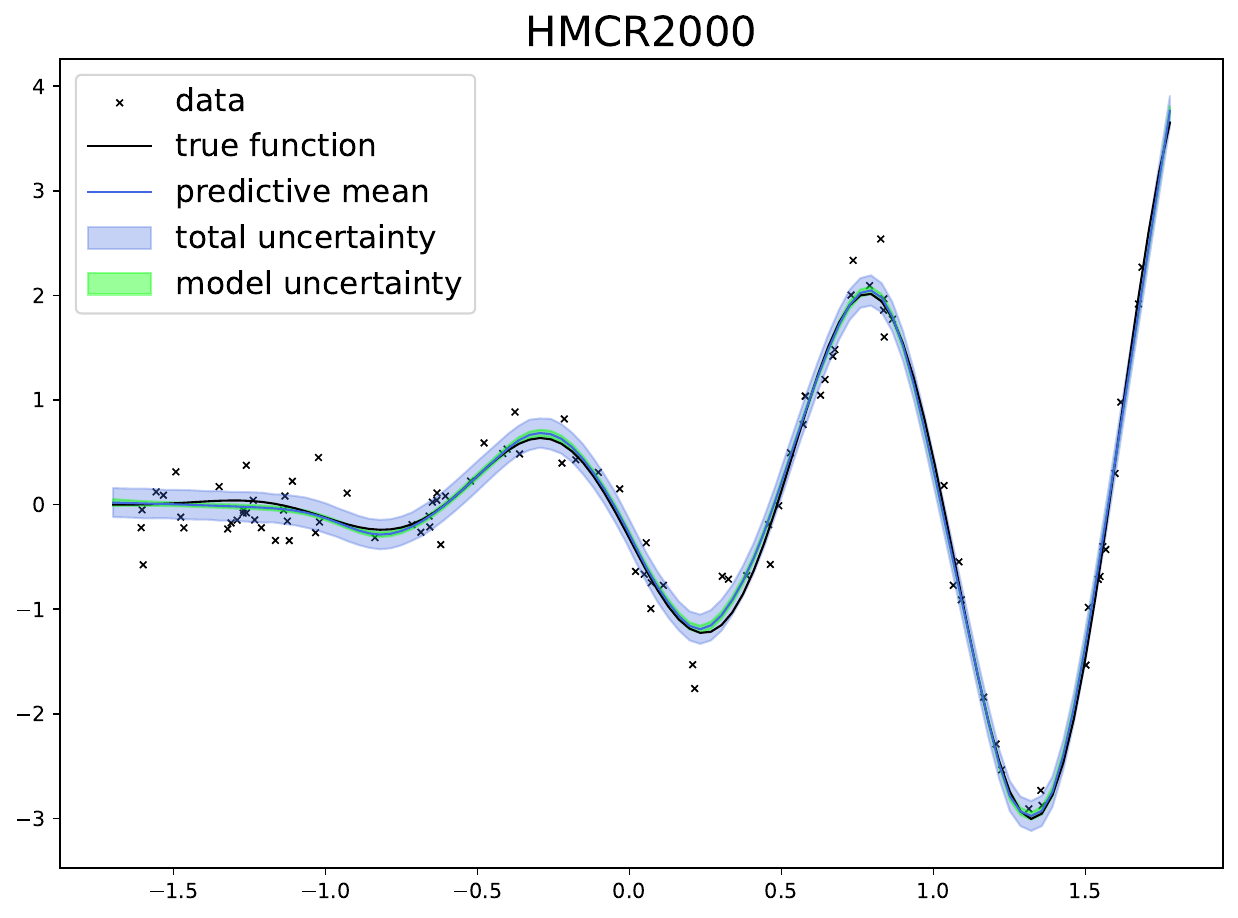}
   \includegraphics[width = 0.49\linewidth]{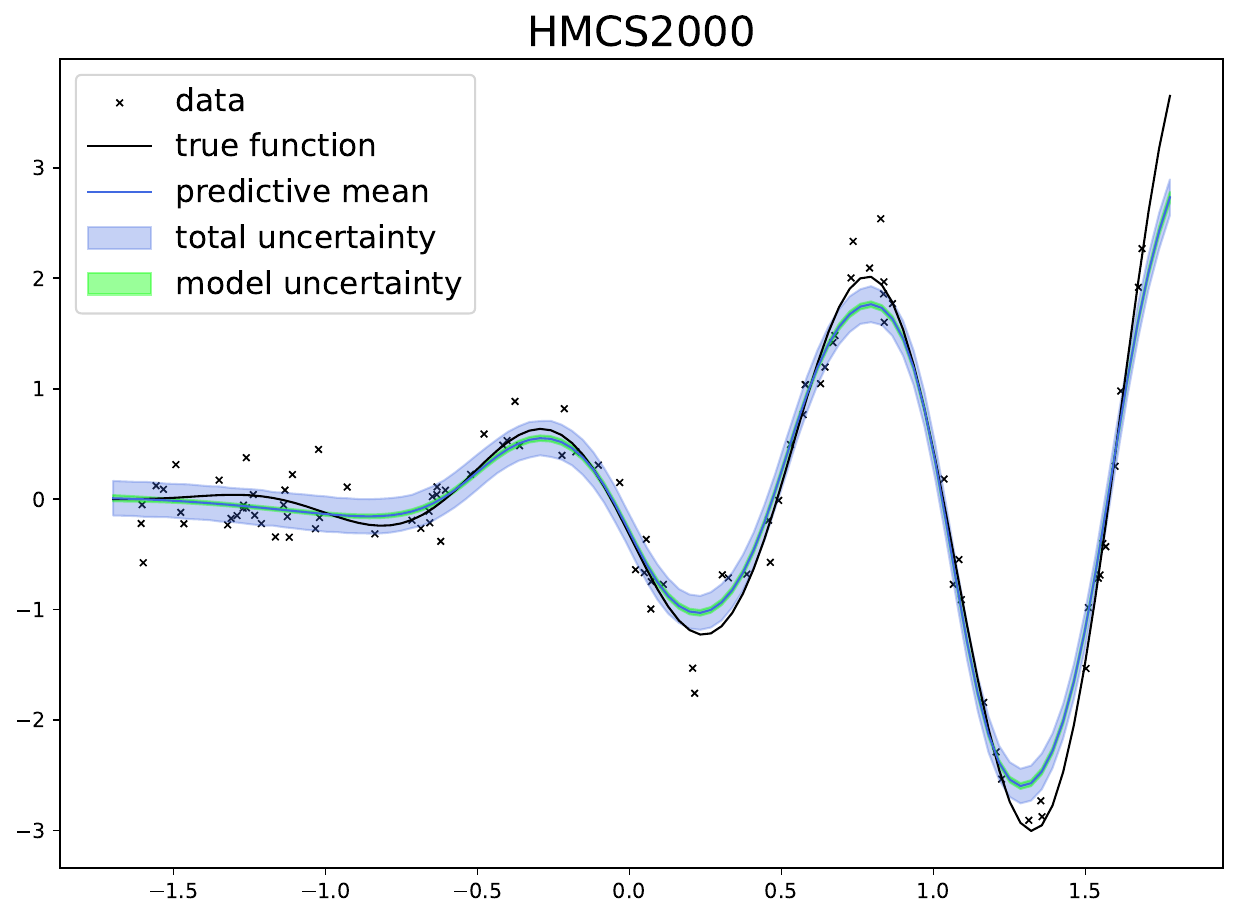}}
\caption{Predictive performance of BNNs as the width increases.}
\label{test}
\end{figure}

We consider a simple synthetic dataset with one-dimensional input  and output: 
\begin{align*}
    \bx & \sim \Unif([0,2]), \; \by = \sin(10\bx)\bx^2 + \bm{\epsilon}, \; \bm{\epsilon} \sim \Norm(0, 0.25).
\end{align*}
The training data $\Data$ consists of $N=500$ observations and the new data for testing $\tilde{\Data}$ consists of $\tilde{N} = 100$ observations.
We first study the performances of mfVIR, mfVIS, HMCR and HMCS with 1 hidden layer and either Gaussian or Student-t priors as the width increases, and illustrate the metrics for $D_1 = 20, 200, 1000$ and $2000$ hidden units by the \cref{onelayermetrics}.  The predictions of the four combinations of activation and inference algorithm with Gaussian priors when $D_1=2000$ are provided on the \cref{2000plots}; similar results were obtained when weights have Student-t distribution, the figures are presented in \cref{appendix:studentt}.
For either choice of priors, performance of the mfVIS dips with the increase in the dimension of the hidden layer; moreover, for $D_1=1000$ and $D_1 = 2000$ its posterior predictive distribution fails to capture the data, and, in fact, degenerates to the prior (\cref{2000plots}).  An explanation of why such behaviour occurs was obtained via the correspondence of Gaussian processes and BNNs. While as the width increases the true posterior of a BNN converges to a NNGP posterior \citep{hron2020exactposteriordistributionswide}, any optimal mean-field Gaussian variational posterior of a BNN with odd (up to a constant offset) Lipschitz activation function converges to the prior predictive distribution of the NNGP \citep{coker2022wide}. In other words, the mean-field variational approximations of wide BNNs with sigmoid activations ignore the data. If one abandons the mean-field assumption and proposes a full-rank variational family, then using variational inference (VI) for wider networks would take at least a hundred times more time than using HMC, which undermines the benefits of using VI. Such degenerate behaviour is not observed with HMC(\cref{2000plots}, but this comes at a significant increase in training time. For wider networks, the HMCR model exhibits a better performance than the HMCS both in terms of accuracy and uncertainty quantification.  In terms of predictive accuracy, HMC is preferred over mfVI in all of the combinations of the activation function and width.
However, in terms of uncertainty quantification, the HMC is inferior to mfVI (with one exception of a BNN with Student-t priors, sigmoid activation and 2000 hidden units). In our experiment, HMC underestimates the uncertainty of the signal much more than VI (\cref{2000plots,onelayermetrics}).
Note that whilst variational inference is often cursed to underestimate the uncertainty\citep{trippe2018overpruningvariationalbayesianneural}, that is not always the case \citep{blundell2015weight,gal2016dropout}.  Markov chain Monte Carlo (MCMC) methods are known to struggle to effectively explore multimodal posteriors \citep{papamarkou2022a,whatabnnposteriorizmailov21a}, and a lack of uncertainty could be a result of poor mixing of the chain. 

\textbf{General summary.} 
In wider networks, the ReLU is preferred over the sigmoid activation for both HMC and mfVI. Crucially, \textbf{when it comes to the mean-field VI the sigmoid activation should only be used when the limited width is suitable for the task at hand}. It is reasonable to suppose that the same could be said about any odd (up to adding a constant) activation function. Further, while the HMC was preferred over the mfVI when looking at accuracy alone, the required computational resources could be an obstacle. Moreover, uncertainty quantification is far from ideal for HMC (CIs are too narrow for the signal); instead, mfVI with the ReLU achieves a good balance between accuracy, UQ, and time, particularly for wider networks.
\subsection{Increasing the Depth of the  Networks} \label{sec:limitdepth}
\begin{figure}[ht]
    \centering
    \subcaptionbox{\centering Metrics of methods as the number of hidden layers increases.\label{depthmetrics} }[.4\textwidth]{ \includegraphics[width=\linewidth]{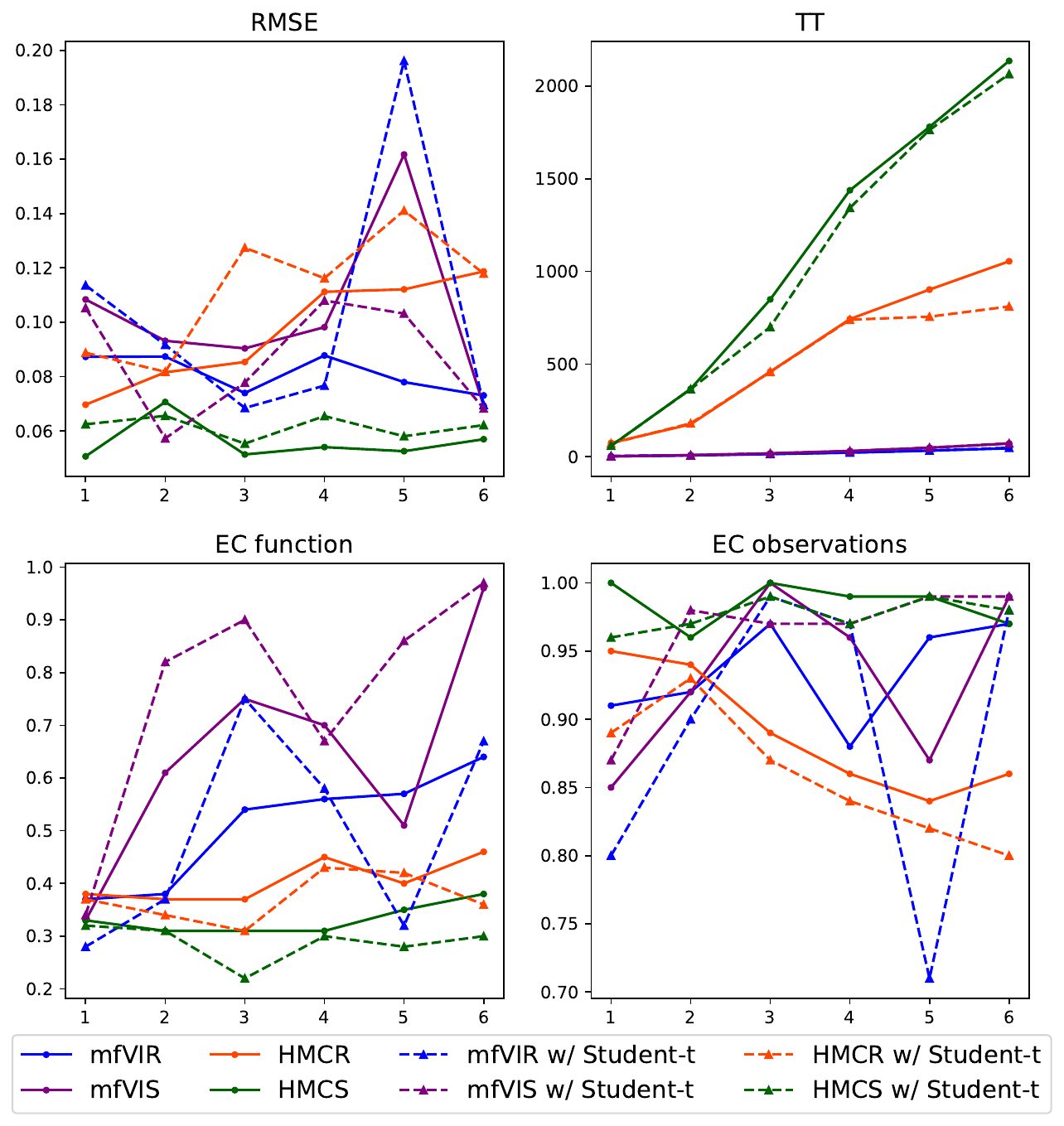}}
    \subcaptionbox{\centering Predictions and uncertainty estimates for each method when $L=6$ and Gaussian priors. \label{depthex}}[.59\textwidth]{ 
    \includegraphics[width=.49\linewidth]{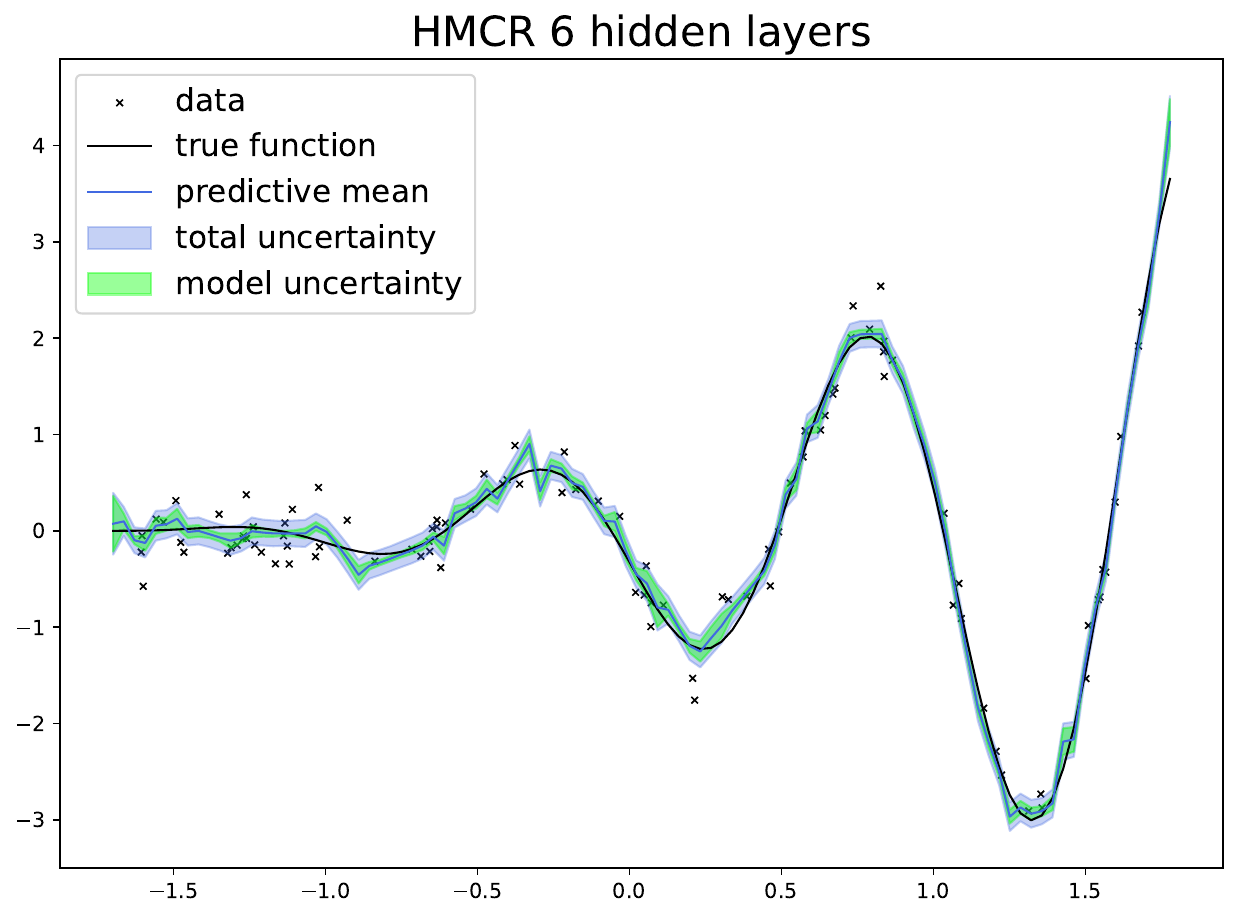}
    \includegraphics[width=.49\linewidth]{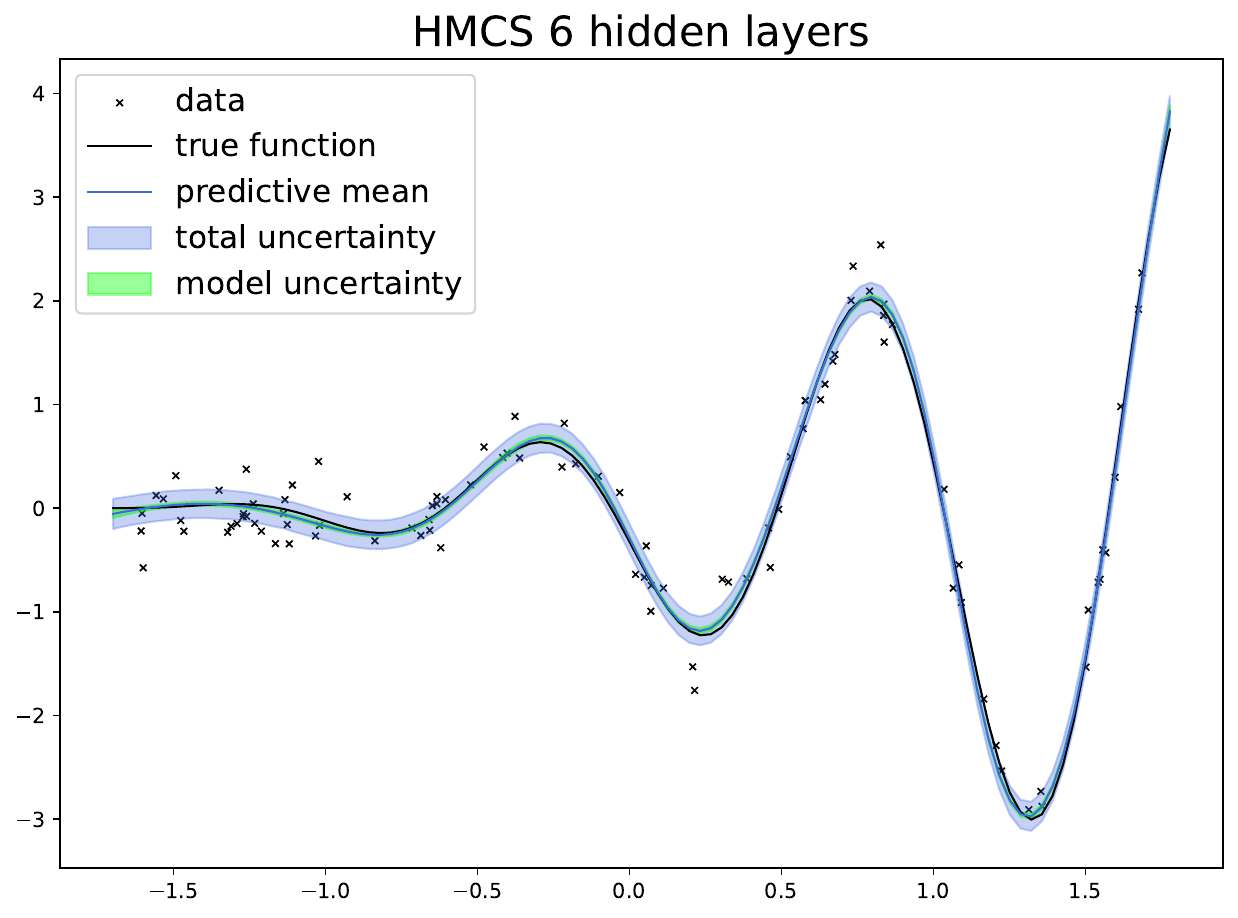}
    \includegraphics[width=.49\linewidth]{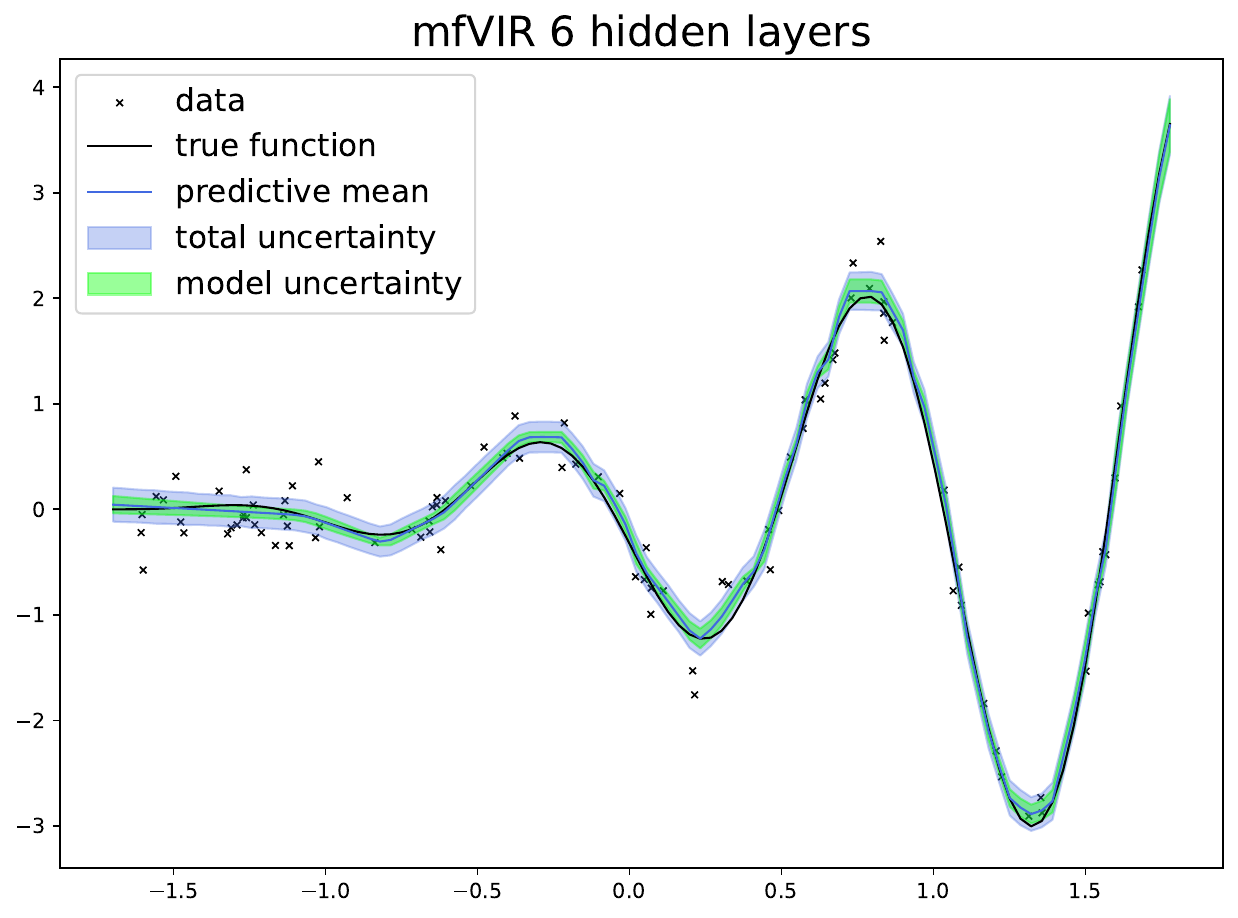}
    \includegraphics[width=.49\linewidth]{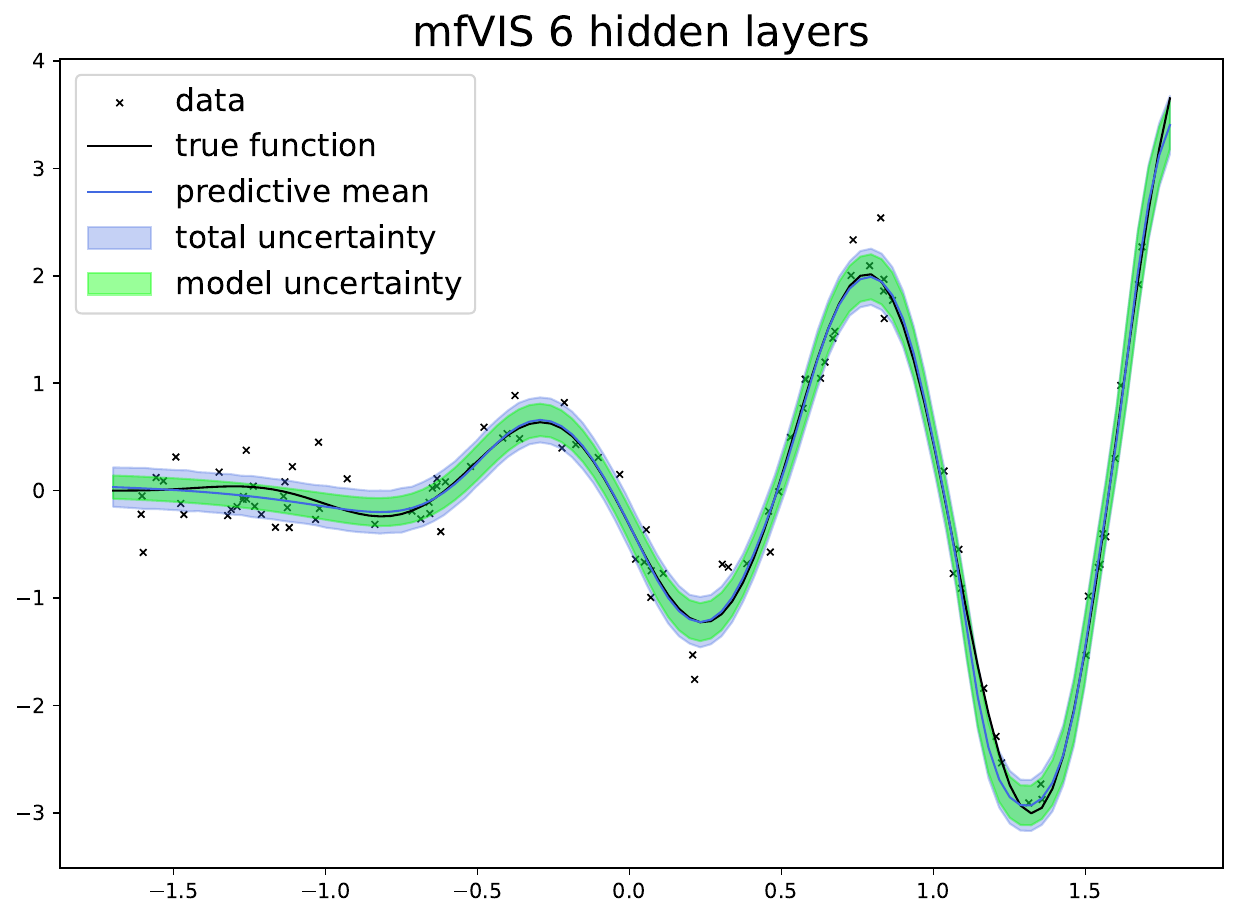}}
    \caption{Prediction performance of BNNs as the depth increases.}
    \label{depthfig}
\end{figure}

Consider the data of \cref{sec:limitwidth} and neural networks defined by \cref{eq:model_for_experiments} with the number of layers $L$ varying from 1 to 6 and a fixed number of hidden units in each layer $D_h =20$.
 \cref{depthmetrics} provides the recorded metrics, and \cref{depthex} illustrates the predictions of the four combinations of activation and inference algorithm with $L=6$ and Gaussian priors (analogous figures for Student-t priors are provided in \cref{appendix:studentt}). 
First, observe that overall both $\RMSE$ and empirical coverage of mfVIR approximations improve with the increase of depth, with one exception of $L=5$ and Student-t priors, when the prediction quality of the network drops drastically. The mfVIS follows a similar pattern, except for the case of $L=5$ and Gaussian priors.  Indeed, 
the approximate posteriors of deep neural networks obtained with the mean-field variational inference were shown to be as flexible as the much richer approximate posteriors of shallower BNNs \citep{farquhar2020liberty}. We do not obtain the same improvement in the prediction quality of models trained with HMC: for either choice of priors, the performance of HMCR falls, whilst the HMCS does not improve as the depth increases. This undesirable behaviour could be a result of the multimodality of distributions in overparametrized models combined with the challenges of MCMC in exploring the high-dimensional space \citep{whatabnnposteriorizmailov21a,papamarkou2022a}.  
Compared to the findings of \cref{sec:limitwidth}, we note that the deeper NNs are less sensitive to the choice of the activation function. It is needless to say that the HMC algorithm scales rather poorly, and as the number of layers changes from $L=1$ to $L=6$, the time needed to train HMCR and HMCS gets more than 15 and 30 times greater, respectively.  We note that for models with more than one hidden layer, training of the network with sigmoid activations takes roughly twice as much time as the network with ReLU. The striking discrepancy in training times could arise due to the difference in the leapfrog integrator step sizes \citep{betancourt2014optimizing}. 

\textbf{General summary.}
 In terms of the training time, HMC becomes less and less feasible with the increase in depth. With the need to explore high-dimensional parameter spaces, multimodality of the posteriors should be kept in mind as an arising challenge for both mfVI and HMC. \textbf{In terms of the balance between accuracy and UQ, the mean-field variational inference with ReLU activation function is able to outperform MCMC with the increase in depth.}

\subsection{Out-of-Distribution Prediction} \label{sec:ood}
While it is not surprising that the accuracy and the quality of uncertainty quantification of any model decreases under a distribution shift, reliable uncertainty estimates that are robust to the out-of-distribution (OOD) data become exceptionally important in safety-critical applications. The challenge is especially intricate since better accuracy and lower calibration error of a certain model on the in-domain data do not imply better accuracy and lower calibration error in the OOD settings \citep{Ovadia2019canyoutrust}. Here, we wish to validate the models' predictive abilities when the test data points come from previously unseen regions of data space. The kind of out-of-distribution data we consider could be described as 'complement-distributions', such data arises in open-set recognition or could be the result of an adversary \citep{farquhar2022what}.  Note that in \cref{sec:stackingexample} as well as in the supplementary, we consider a much milder example with 'related-distributions' data. 
We split the training data used in \cref{sec:limitwidth,sec:limitdepth} into the train and test data covering complement regions of the function.  

\begin{figure}[t]
    \centering
    \subcaptionbox{\centering Out-of-sample metrics of all the methods compared.\label{oodonelayermetrics} }[.4\textwidth]{ \includegraphics[width=\linewidth]{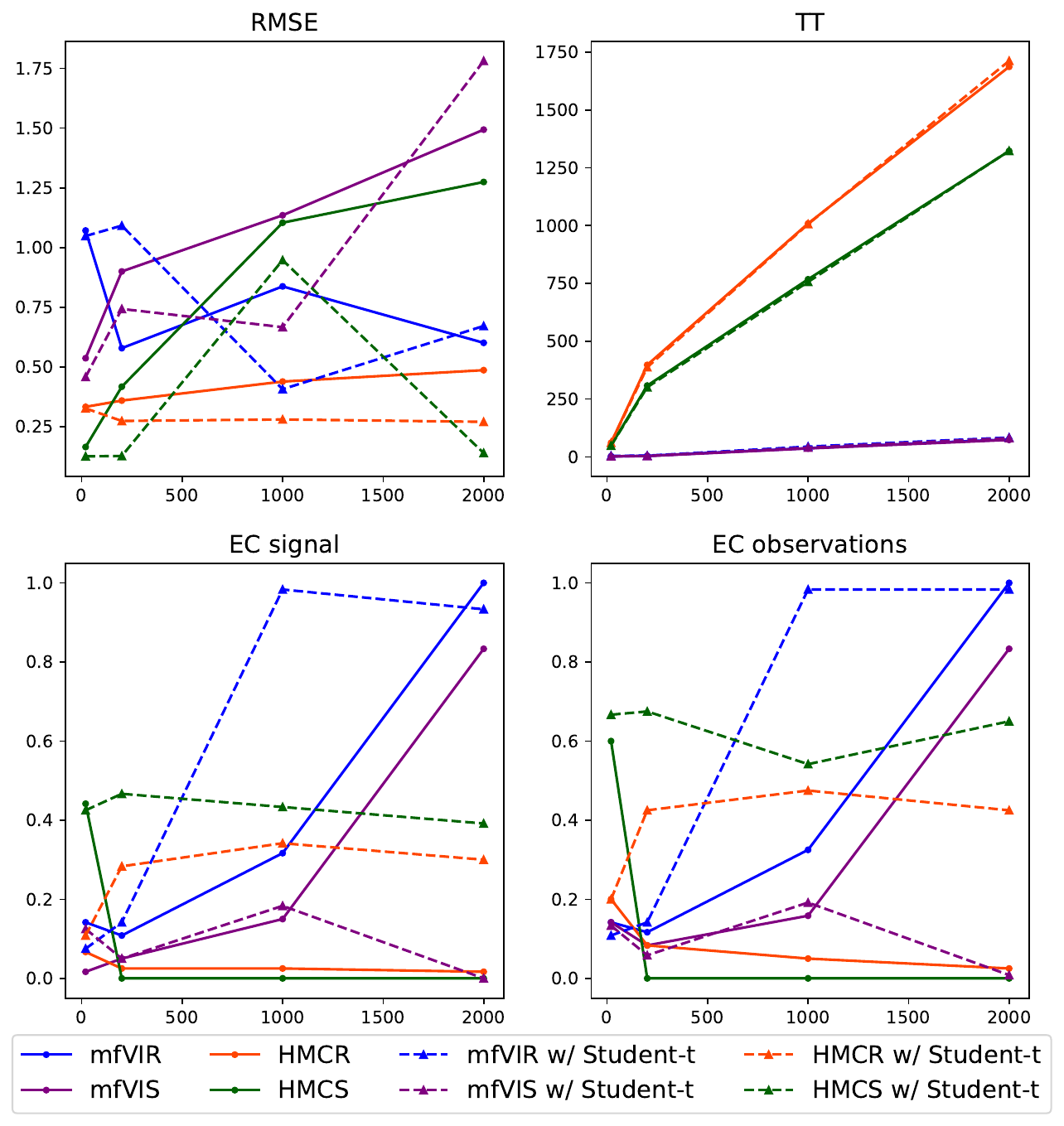}}
    \subcaptionbox{\centering Within-the sample (left) and out-of-sample (right) predictions and uncertainty estimates  for $D_1=200$, ReLU activation and Gaussian priors.\label{mfvir200hmcr200} }[.59\textwidth]{ \includegraphics[width=0.49\linewidth]{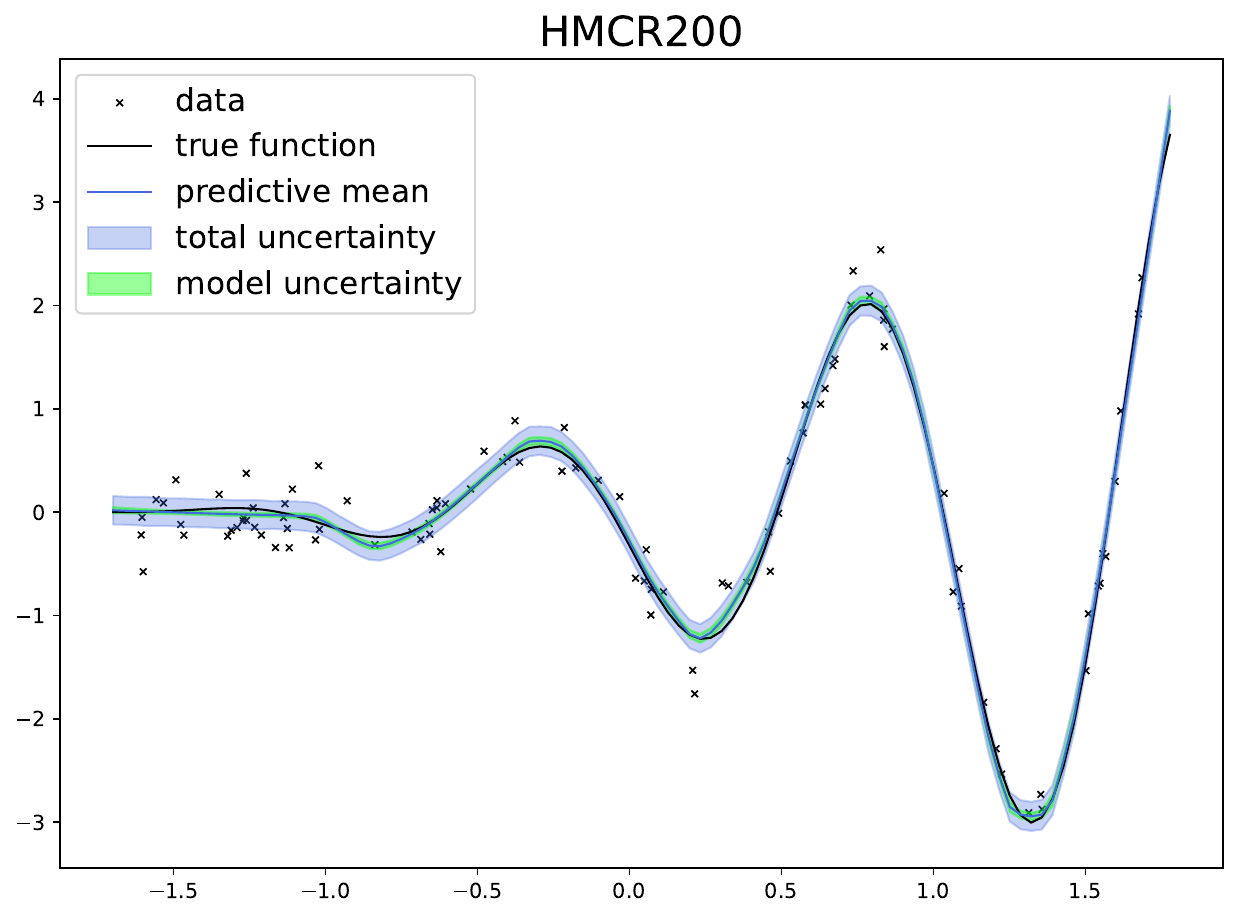}
    \includegraphics[width=0.49\linewidth]{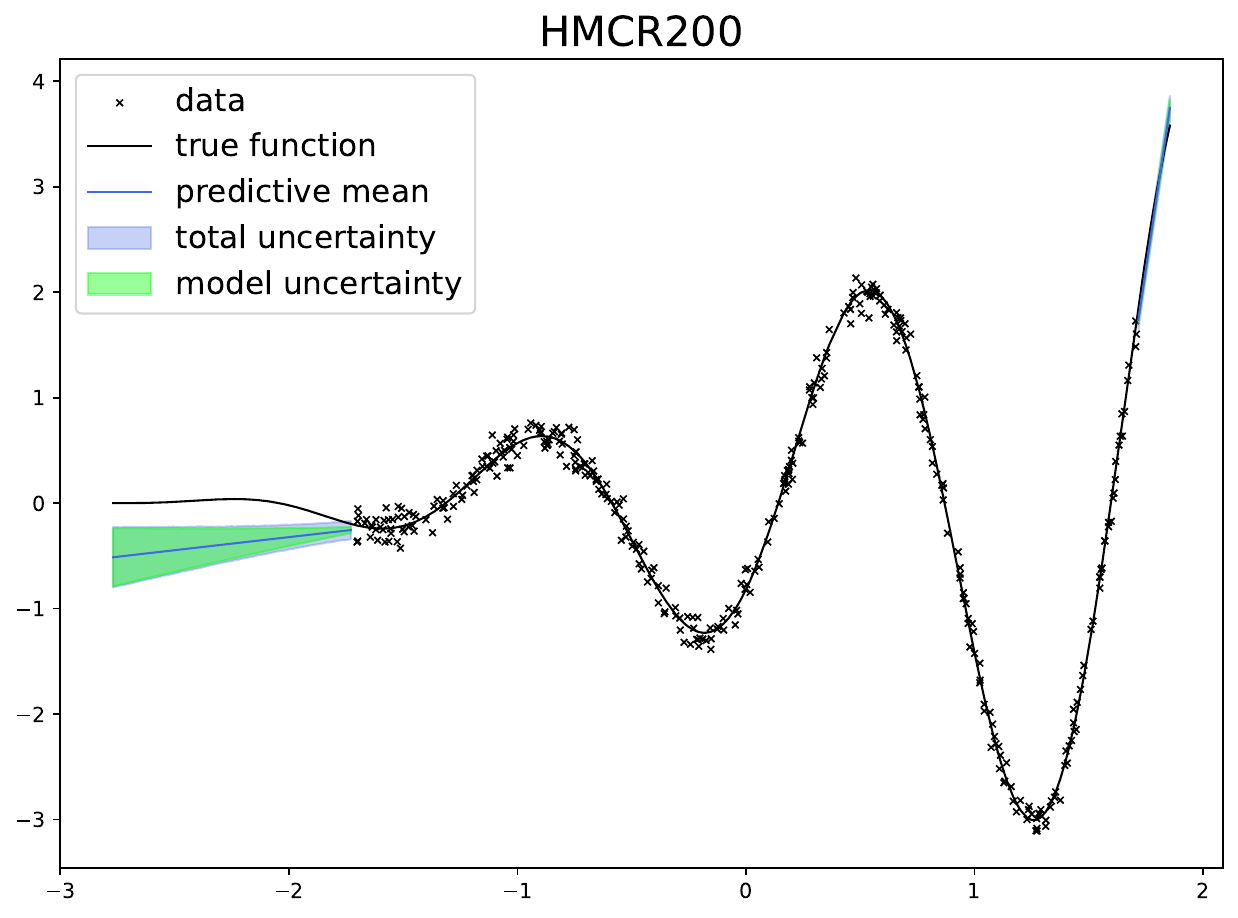}
    \includegraphics[width=0.49\linewidth]{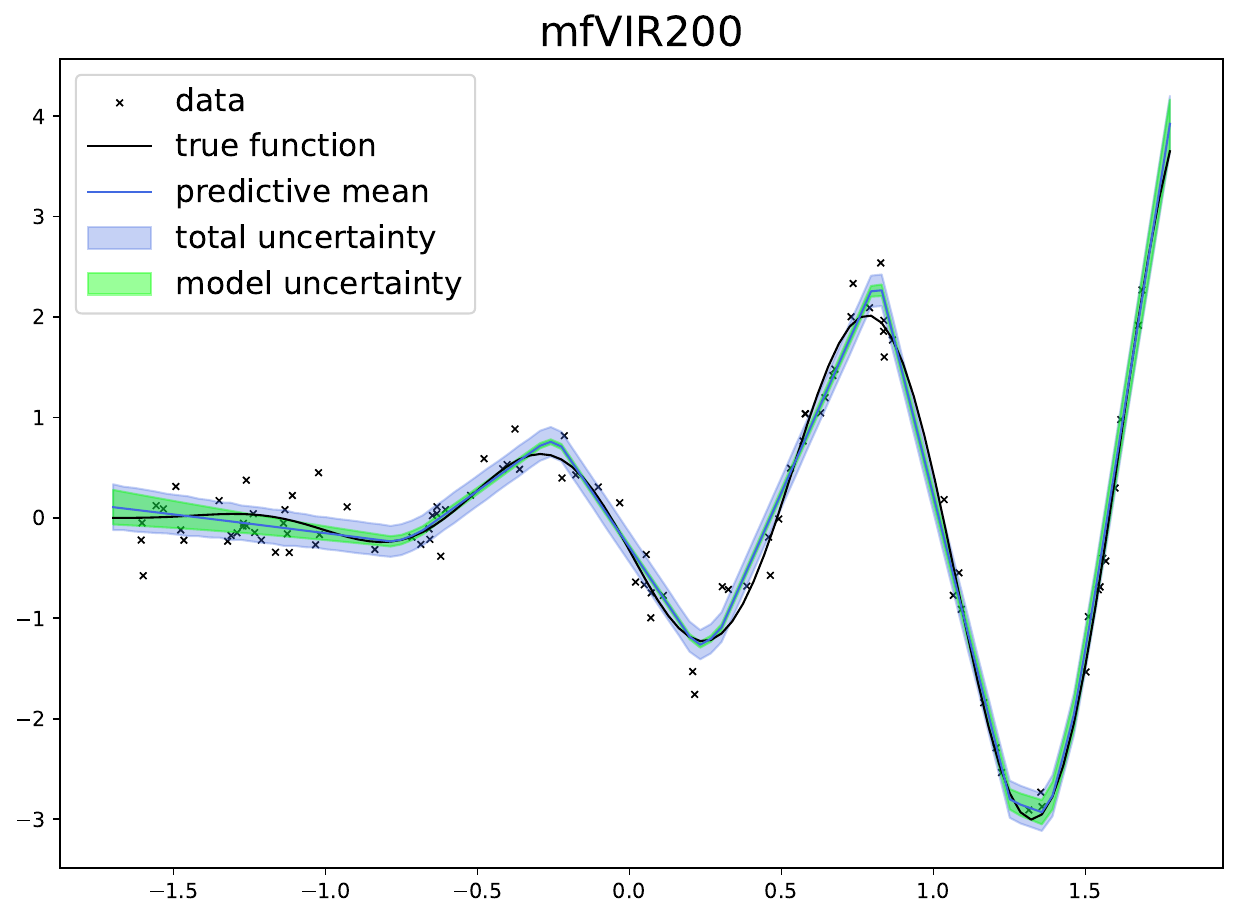}
    \includegraphics[width=0.49\linewidth]{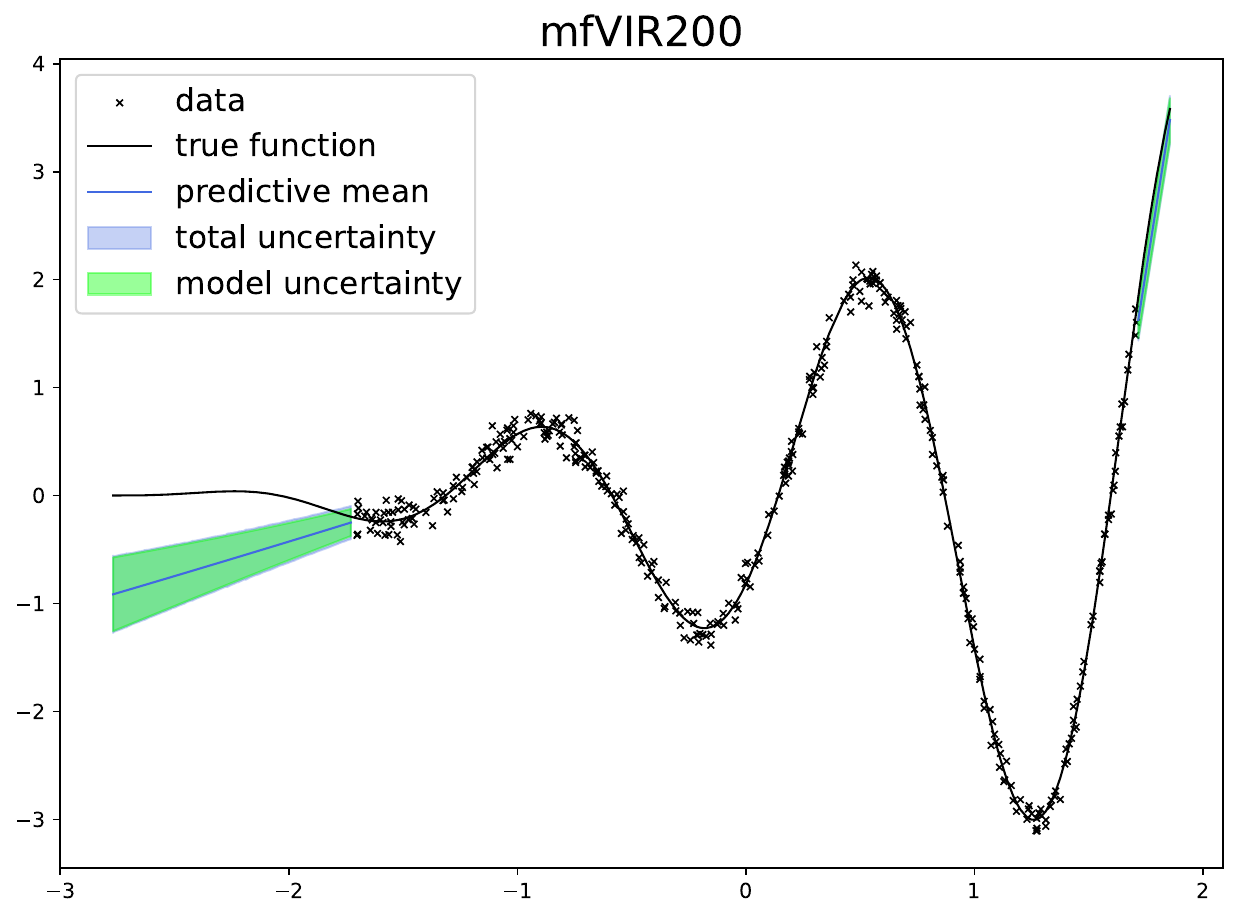}}
    \caption{Out-of-distribution prediction for the 'complement-distributions' data.}
    \label{oodonelayerfigures}
\end{figure}

Specifically, $ \Data = \Data_c \sqcup \tilde{\Data}_c$, the observed data $\Data_c$ consists of $N=370$, the new data $\tilde{\Data}_c$ consists of $\tilde{N} = 130$ and the observed and the new data are disjoint (see \cref{oodonelayerfigures}): 
\begin{align*}
    \Data_c & = \{(x_n, y_n) \; | \; x_n \in [-1.7, 1.7]\},\\
    \tilde{\Data}_c & = \{(x_n, y_n) \; | \; x_n \in [-2.8, -1.7) \cup (1.7, 1.9)]\}.
\end{align*}
Strictly speaking, we do not expect any model to be robust to such an extreme case and, mainly, want to asses and better understand the quality of the uncertainty estimates. In this experiment, we are hoping that the relationship between the distributions of the observed and the new data makes this challenge somewhat tractable.
 On \cref{oodonelayermetrics} we illustrate the metrics for $D_1 = 20, 200, 1000$ and $2000$ hidden units; \cref{mfvir200hmcr200} compares non-OOD and OOD predictions obtained by the BNNs with ReLU activation, Gaussian priors and $D_1=200$.
The poor performance of the mfVIS, especially for wider networks, is not surprising, however, we notice that for wide networks HMCS with Gaussian priors suffers from much higher $\RMSE$ than HMCS with Student-t priors and mfVIR and HMCR with either choice of priors. And while HMCR has a lower $\RMSE$ than any model trained with mean-field VI, the ability of HMC to capture the uncertainty deteriorates, and it becomes overconfident.  
Whilst HMCR200 and mfVIR200 do not show any of the expected increase in the uncertainty, on certain regions both methods are able to provide accurate predictive mean (see \cref{mfvir200hmcr200} for examples with Gaussian priors, the right-hand side region of the function, where $x>1.5$). Finally, as the width of the network increases, mfVIR outperforms all of the methods. 

\textbf{General summary.} In terms of the accuracy alone, the HMC with ReLU is more robust to the out-of-distribution data, however, that comes with the largest computational costs among all the models. We already saw in \cref{sec:limitwidth} that uncertainty quantification with HMC degrades with increasing width. In OOD settings, this becomes even more extreme, with very overconfident predictions that do not cover the truth (an empirical coverage of almost zero).
\textbf{Finally, with the increase in depth, in the extreme OOD settings, the mfVI with ReLU becomes almost as accurate as HMC with ReLU and provides better UQ at a much lower cost.} 

\section{Bayesian Model Averaging and Stacking} \label{sec:stackingbig}
\subsection{Predictive Methods for Model Assessment} \label{sec:elpd}
When considering synthetic datasets, we can choose a desired metric and sample any number of data points, so that evaluation of the model’s performance becomes trivial. For example, in \cref{sec:ood} we have specifically created an extreme case when the training data $\Data_c$ and the new data $\tilde{\Data}_c$ were covering disjoint regions of the true function. In reality, the new previously unseen data is not available, and one can only estimate the expected out-of-sample predictive performance.
 Suppose that we only observe $\Data$, the unseen observations $\tilde{\Data}$ are generated by $p_t(\tilde{\Data})$, and we wish to be able to assess the generalization ability of the model without having access to the test data. To keep the notation simple, we omit the dependency on $\bx$ and $\Tilde{\bx}$ when writing down the posteriors in this section. Given a new data point $\tilde{y}_n$, the log score $\log p(\tilde{y}_n |  \Data)$ is one of the most common utility functions used in measuring the quality of the predictive distribution. The log score benefits from being a local and proper scoring rule \citep{VehtaryOjanen2012asurvey}. Then, the expected log pointwise predictive density for a new dataset  serves as a measure of the predictive accuracy of a given model:
\begin{align*}
    \text{elpd} = \sum_{n=1}^{\Tilde{N}} \int p_t(\Tilde{\mathrm{D}}_n) \log p(\tilde{y}_n |  \Data) \dif \Tilde{\mathrm{D}}_n,
\end{align*}
where $p(\tilde{y}_n |  \Data)$ is model's posterior predictive distribution. In the absence of $\tilde{\Data}$, one might obtain an estimate of the expected log pointwise predictive density by re-using the observed $\Data$. Here, we review the approach that employs leave-one-out cross-validation (LOO-CV), which can be seen as a natural framework for assessing the model's predictive performance \citep{VehtariGelmanGabry_2016}. 

To obtain the Bayesian leave-one-out cross-validation (LOO-CV) estimate of the expected utility $\loo$ and avoid re-fitting the model $N$ times, one could use importance sampling. However, the classical importance weights would have a large variance, and the obtained estimates would be noisy. Recently, the problem was solved with Pareto smoothed importance sampling (PSIS), which allows evaluating the LOO-CV expected utility in a reliable yet efficient way \citep{VehtariGelmanGabry_2016}:
\begin{align}
   \loo  &= \sum_{n=1}^N p(y_n | x_{n}, \Data_{-n}) = \sum_{n}^N \log \left(\frac{\sum_{s=1}^S r_i^s p (y_n | \btheta^s)}{\sum_{s=1}^Sr_i^s}\right ),  \label{eq:elpdloo}
\end{align}
where $r_i^s$ are the smoothed importance weights, which benefit from smaller variance than the classical weights. We refer to the individual logarithms in the sum as  $\widehat{\text{elpd}}_{\text{loo}, n}$. The advantage of PSIS is that the estimated shape parameter of the Pareto distribution provides a diagnostic of the reliability of the resulting expected utility. Although methods of model selection which reuse the data can be vulnerable to overfitting when the size of the dataset is too small and/or the data is sparse, it is (relatively) safe to use cross-validation to compare a small number of models and given a large enough dataset \citep{Vehtari19a}. 
In \cref{appendix:predictivemodelassesment}, we implement $\loo$ in the empirical experiment, where we additionally consider posterior predictive checks (PPC) and an alternative to the LOO-CV approach of estimating the expected log pointwise utility.

\subsection{Alternatives to Classical Bayesian Model Averaging}\label{sec:stacking}
Let $\mathcal{M} = \{M_1, \ldots, M_K\}$ be a collection of models and denote the parameters of each of the $M_k$ as $\btheta_k$. 
The assumptions one has on the prediction task and on $\mathcal{M}$ with respect to the true data-generating process can be categorized into three scenarios: $\mathcal{M}$-closed, $\mathcal{M}$-open and $\mathcal{M}$-complete.  If $M_k \in \mathcal{M}$ for some $k$ recovers the true data generating process, then we are in the $\mathcal{M}$-closed case. The task is $\mathcal{M}$-complete if there exists a true model but it is not included in  $\mathcal{M}$ (e.g. for computational reasons). Finally, we are in the $\mathcal{M}$-open scenario when the true model is not in $\mathcal{M}$ and the data-generating mechanism cannot be conceptually formalized to provide an explicit model \citep{VehtaryOjanen2012asurvey}. The Bayesian framework allows to define the probabilities over the model space, 
and for the $\mathcal{M}$-closed case, classical Bayesian Model Averaging (BMA) would give optimal performance. The BMA solution provides an averaged predictive posterior as  \citep{hoeting1999bayesian}
\begin{align}
    p (\tilde{\by} \mid \Data) &= \sum_{k=1}^K p(\tilde{\by} \mid \Data, M_k) p (M_k\mid \Data), \label{eq:BMA} \\
    \text{ where  }  & p(M_k \mid \Data ) \propto p(\Data \mid M_k) p (M_k).
\end{align}
However, in the $\mathcal{M}$-open and $\mathcal{M}$-complete prediction tasks, BMA is not appropriate as it gives a strong preference to a single model and so assumes that this particular model is the true one.
Now, if we replace the weights $p(M_k|\Data)$ with the products of Bayesian LOO-CV densities $\prod_{n=1}^N p(y_n \mid x_n, \Data_{-n}, M_k)$, we arrive at pseudo-Bayesian model averaging (pseudo-BMA). In other words, the weights $w_k$ of pseudo-BMA are proportional to the estimated log pointwise predictive density $\exp(\loo^{k})$ introduced in \cref{sec:elpd}. 
One could further correct each $\loo$ estimate of \cref{eq:elpdloo} by the standard errors and obtain
\begin{align*}
    w_k& = \frac{\exp(\loo^{k, \text{reg}})}{\sum_{k=1}^K \exp(\loo^{k, \text{reg}})}, \\
    \loo^{k, \text{reg}}& = \loo^k - \frac{1}{2}\sqrt{\sum_{n=1}^N \left (\widehat{\text{elpd}}_{\text{loo}, n}^k - \frac{\loo^k}{N}\right)^2 },
\end{align*}
where for each model $M_k$ we find $\loo^{k, \text{reg}}$ by utilizing a log-normal approximation.
Fortunately, we have already seen that these densities can be efficiently estimated with PSIS.

An alternative way to obtain the averaged predictive posterior given the set of $p(\tilde{\by} \mid \Data, M_k)$ is to employ the stacking approach \citep{Yao_2018}. 
Define the set $S^K = \{\bw \in [0,1]^K | \sum_{k}^K w_k = 1\}$, then the stacking weights are found as the optimal (according to the logarithmic score) solution of the following problem
\begin{align*}
   \bm{w} & = \max_{ \bm{w} \in S^K} \frac{1}{N}\sum_{n = 1}^{N}\log \sum_{k=1}^{K} w_k p(y_n \mid \Data_{-n}, M_k), \\
   & = \max_{ \bm{w} \in S^K} \frac{1}{N}\sum_{n = 1}^{N}\log \sum_{k=1}^{K} w_k \left (\frac{\sum_{s=1}^Sr_i^s p (y_n \mid \btheta_k^s, M_k)}{\sum_{s=1}^Sr_i^s} \right), 
\end{align*}
where a PSIS estimate of the predictive LOO-CV density is used, and  $r_i^s$ are the smoothed (truncated) importance weights. 

Finally, we recall that deep ensembles of classical non-Bayesian NNs \citep{LakshminarayananDeepEns} behave similarly to Bayesian model averages, and both lead to solutions strongly favouring one single model \citep{wilson2020bayesian}. In contrast, the ensembles of BNN posteriors in  \cref{eq:BMA} with $p (M_k\mid \Data)= K^{-1}$ can be seen as a trivial case of BMA, which combines models and does not give preference to a single solution. Alternatively, when implementing variational inference and combining BNNs, the analogy can be drawn with the simplified version of adaptive variational Bayes, which combines variational posteriors with certain weights and, under certain conditions, attains optimal contraction rates \citep{ohn2024adaptive}.

\subsection{Ensembles and Averages} \label{sec:stackingexample}
We compare three model averaging methodologies: deep ensembles of Bayesian neural networks, stacking and pseudo-BMA based on PSIS-LOO \citep{Yao_2018}. We do not consider the Bayesian Bootstrap (BB) \citep{rubin1981bayesian} motivated by the recent observation that in the settings of modern neural networks deep ensembles of non-Bayesian NNs and BB are equivalent, and both are often misspecified \citep{WuWillianson2024}. 
Combining several estimates of BNNs can be effective not only when predictions are coming from different models, but also when dealing with several predictions obtained by the same model \citep{ohn2024adaptive}. This is of particular use for multimodal posteriors arising in BNNs, where different modes could be explored by random initializations \citep{Yao_2018}. Additionally, recall that the ELBO, the objective of variational inference, is a non-convex function, so that the optimum is only local and depends on the starting point. 
We note that combining models trained with HMC and VI would be meaningless for several reasons. First of all, training a set of HMC models becomes rather expensive: for instance, training the HMCR20 once takes the same amount of time as 35 trainings of mfVIR20. Second, the estimates of the log pointwise predictive densities (provided in \cref{appendix:predictivemodelassesment}) for HMC and VI have different scales and are not easily compared; in this case, the result of averaging HMC and VI would be equivalent to classical BMA. 

\begin{figure}[t]
    \centering
     \includegraphics[width=0.9\linewidth]{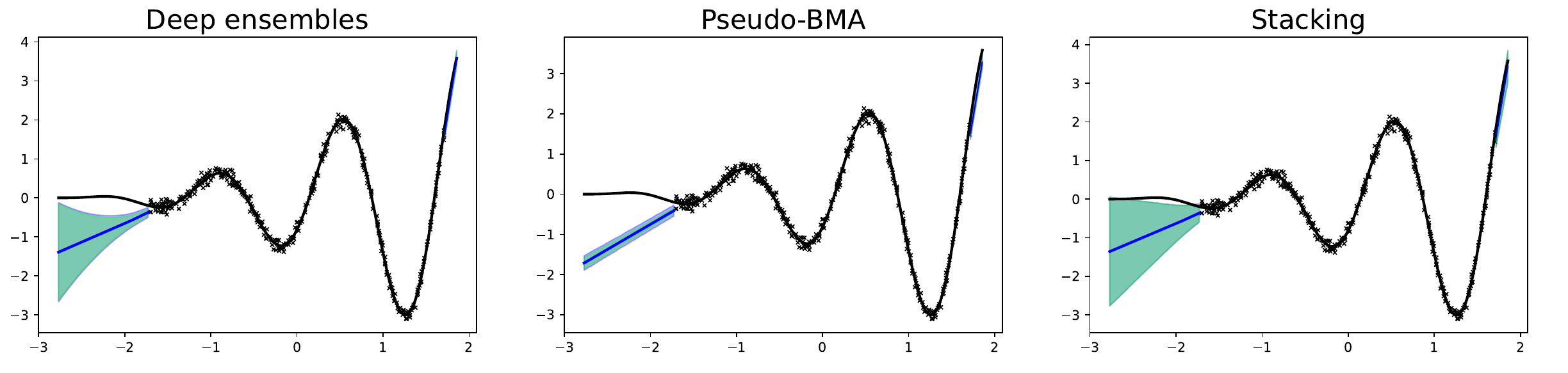}
    \includegraphics[width=0.9\linewidth]{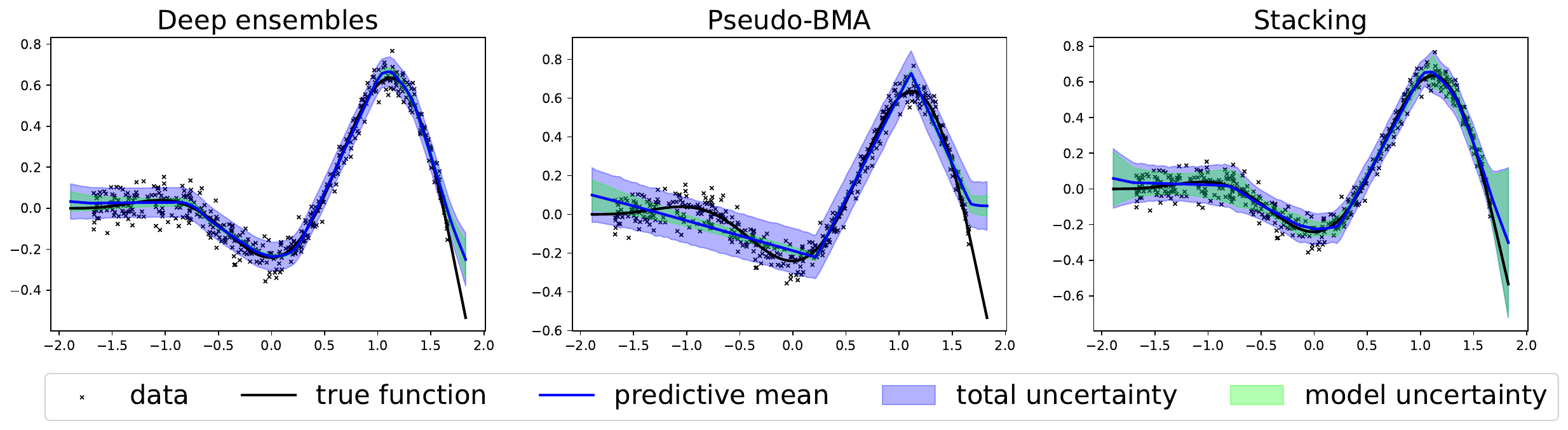}
    \caption{Predictions obtained by ensembling, stacking and pseudo-BMA when applied to mfVIR20 with Gaussian priors in the 'complement-distributions' (top)  and 'related-distributions' (bottom) OOD tasks. Pseudo-BMA is worse than the other methodologies, and stacking provides improvements over DE in uncertainty quantification.}
    \label{fig:OODstack}
\end{figure}

Now consider the mfVIR20 model with Gaussian priors and the 'complement-distributions' data of \cref{sec:ood}. We choose 10 random initialization points, obtain 10 posterior predictive distributions and compute estimated expected log pointwise predictive densities. We then construct ensemble, pseudo-BMA and stacking approximations; the results are illustrated \cref{fig:OODstack}. Ensembling and stacking 
are superior to pseudo-BMA, which has worse accuracy and fails to capture any uncertainty. Similar results for the mfVIR20 model with Student-t priors are provided in \cref{appendix:studentt}. 

%



Given the nature of the test data we use, the predictions as well as the $\loo$ estimates may be unreliable. Thus, 
we consider a simpler data-generating mechanism in which test data comes from a slightly broader region; such a scenario could be called an OOD task with 'related-distributions' \citep{farquhar2022what}. 
Specifically, the data are generated as follows:
\begin{align*}
\bx &\sim \Unif([0,1]),\quad \by = \sin(10\bx)\bx^2  + \bm{\epsilon}, \quad \bm{\epsilon} \sim 0.05 \Norm(0, 1).
\end{align*}
The data for training $\Data_r$ and the testing $\tilde{\Data}_r$ consist of $N=450$ and $\tilde{N} = 50$ observations, respectively, where $\tilde{\Data}_r$ comes from the broader region than $\Data_r$, i.e. $ (\min_{n=1 \ldots N} (x_n), \max_{n=1\ldots N}(x_n)) \subsetneq (\min_{n=1 \ldots \tilde{N}} (\tilde{x}_n), \max_{n=1 \ldots \tilde{N}} (\tilde{x}_n))$. For 10 posterior predictive distributions of mfVIR20 with Gaussian priors (results for Student-t priors are provided in \cref{appendix:studentt}), we compare ensembling, pseudo-BMA and stacking in \cref{fig:OODstack} (similar results with having Student-t priors are presented in the supplementary, and in \cref{appendix:deepernetworks} we additionally provide the results of ensembling and averaging in the deeper networks.). Whilst the total uncertainty estimates of pseudo-BMA are, somewhat, adequate, the model uncertainty is underestimated. Both stacking and deep ensembles lead to improved predictive performance and uncertainty quantification, with stacking showing some better gains compared to deep ensembles (see e.g. improved coverage of stacking on the right-hand side of \cref{fig:OODstack}).

\textbf{General summary.} \textbf{We observe that, similar to BMA, the pseudo-BMA is not preferable in $\mathcal{M}$-open and $\mathcal{M}$-complete settings.} Namely, in 'complement-distributions' and 'related-distributions'  experiments,  pseudo-BMA was confirmed to be inferior to stacking and ensembles of BNNs both in terms of the predictive accuracy and uncertainty quantification. \textbf{Stacking and ensembles of BNNs performed comparable to each other and provided an improvement, with modest gains for stacking, which is especially significant in terms of uncertainty quantification in the OOD setting.}
\subsection{The Rocket Booster Simulation} \label{sec:NASA}
We further evaluate the empirical performance of Bayesian model averaging techniques in BNNs by considering the data simulated by NASA when designing a novel rocket booster called Langley Glide-Back Booster (LGBB) \citep{gramacy2009bayesian,rogers2003automated}. The data comprises 3167 observations, and the goal is to model the lift force as a function of the speed of the vehicle, the angle of attack $\alpha$, and the sideslip angle $\beta$. Speed ranges from Mach 0 to Mach 6, the angle $\alpha$ lies in between $-5$ and $30$ degrees and the sideslip angle  $\beta$ takes values $0,0.5,1,2,3,4$.

\begin{figure}[t!]
    \centering
    \includegraphics[clip, trim=0.0cm 3cm 0.0cm 0.0cm, width=0.8\linewidth]{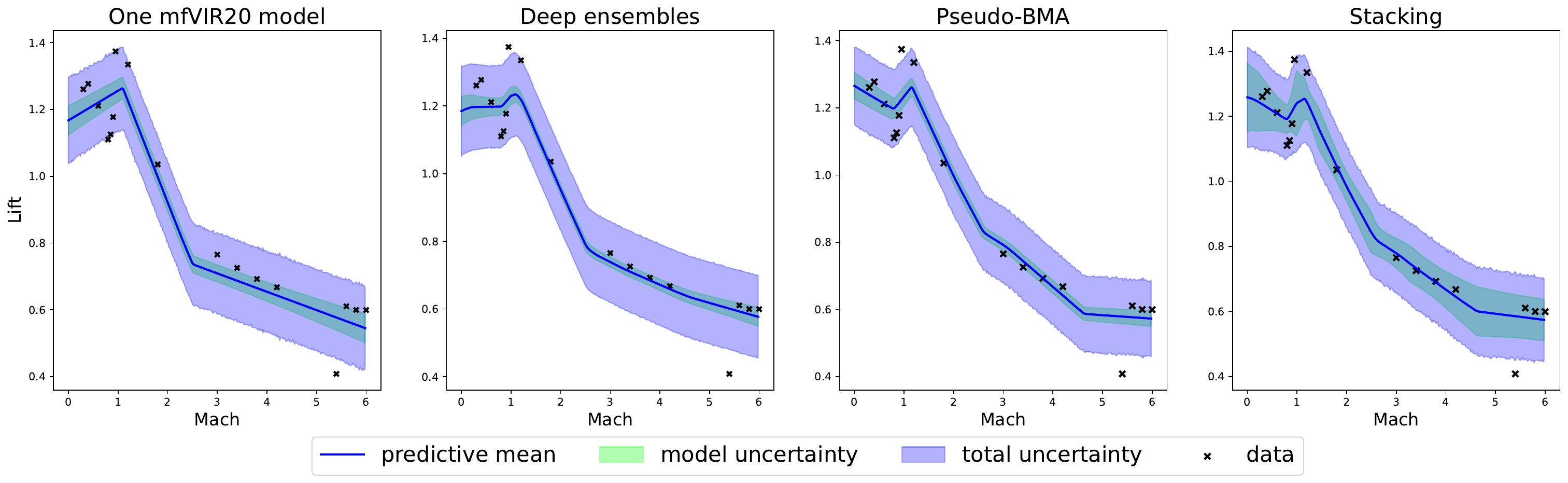}
    \includegraphics[width=0.8\linewidth]{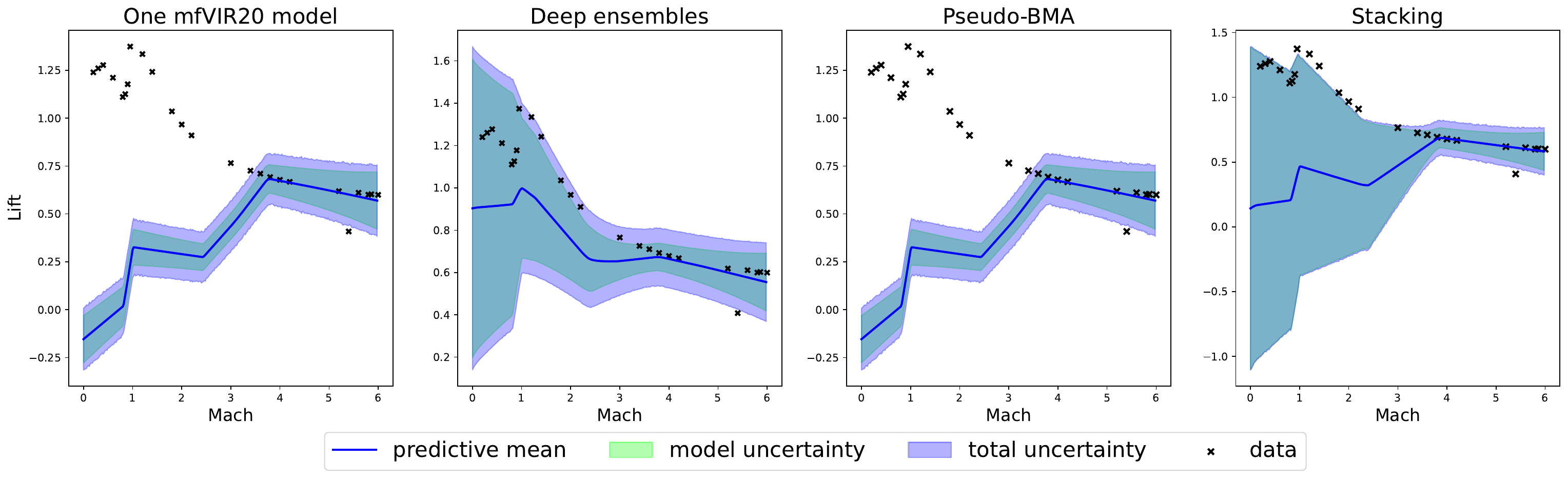}
    \caption{Slices of predicted lift obtained by a single model and by ensembling, stacking and pseudo-BMA when applied to mfVIR20 with Gaussian priors in the non-OOD (top) and 'complement-distributions' OOD (bottom) tasks.}
    \label{fig:nasa_slice}
\end{figure}

We begin with the non-OOD scenario, consider a one-layer BNN with ReLU activation, Gaussian prior and $D_1 = 20, 100$ or $200$ and train our model with mfVI on $80\%$ of the data. 
\cref{fig:nasa_slice} displays a slice of predicting lift, with Mach (speed) on the x-axis, $\alpha$ fixed to 25 and $\beta$ to 4; it compares predictions of a single mfVIR20 model to the results of combining 10 predictive distributions via deep ensembles, pseudo-BMA and stacking.  Figures for slices of predicting lift obtained for models with $100$ and $200$ hidden units are provided in \cref{appendix:nasadata}. \cref{fig:metricsnasa} compares the metrics obtained by different approaches as the BNN gets wider. 
Stacking provides modest gains in accuracy, and while coverage is comparable across all methods, the slightly wider model uncertainty visible in the top row \cref{fig:nasa_slice} for stacking seems more reasonable given the observed data (crosses). 

Further, we construct the 'complement-distributions' OOD scenario: the data taken for training consists of all datapoints but those whose sideslip angle $\beta$ equals 4, the data on which we then test the model consists of the remaining datapoints for which $\beta=4$. 
Similar to the non-OOD scenario, \cref{fig:nasa_slice} (bottom) compares slices of predicting lift obtained by a single mfVIR20 model to the predictions of deep ensembles, pseudo-BMA and stacking (figures for models with $100$ and $200$ hidden units are provided in \cref{appendix:nasadata}); 
this figure particularly highlights how a single model and pseudo-BMA can lead to incorrect, overconfident predictions in OOD settings. 
Unlike the non-OOD case, where performance is stable across the network's width, it varies substantially for OOD (left versus right side of \cref{fig:nasa_slice}), especially for pseudo-BMA and stacking, which show improvements in both accuracy and uncertainty as the width increases. 
\begin{figure}
    \centering   \includegraphics[clip, trim=0.0cm 2.cm 0.0cm 0.0cm, width=0.45\linewidth]{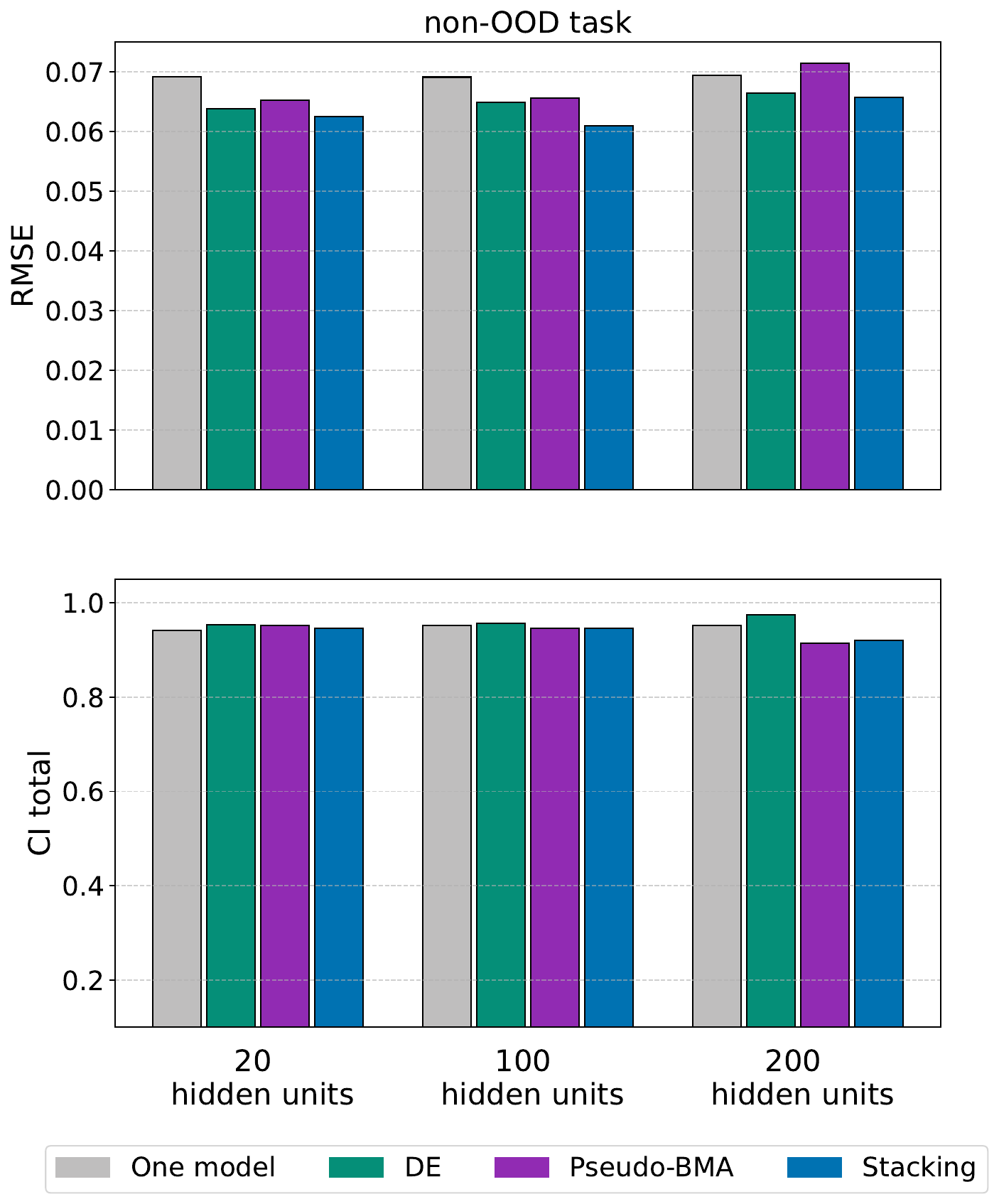}
\includegraphics[clip, trim=0.0cm 2.cm 0.0cm 0.0cm, width=0.45\linewidth]{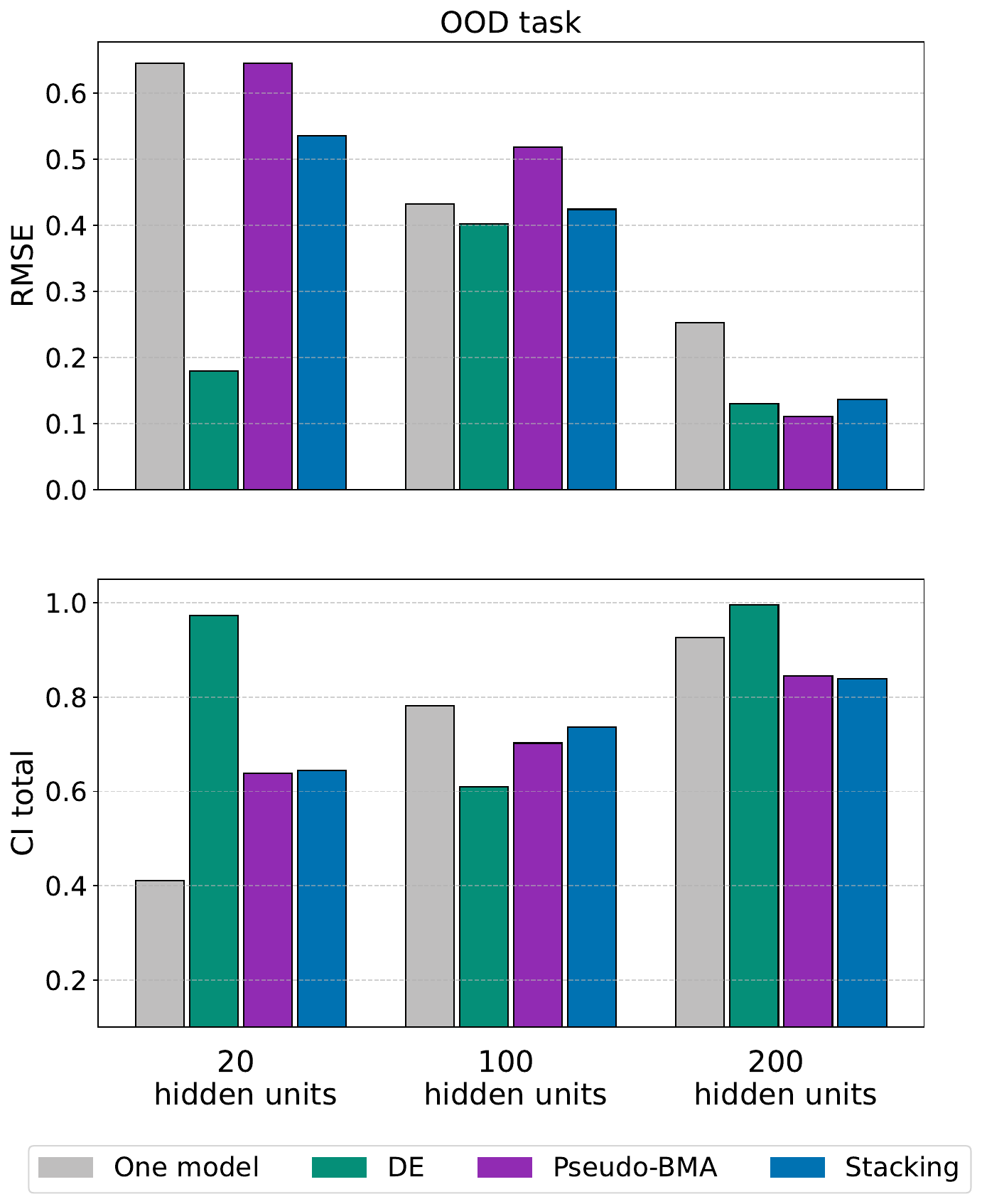}
        \includegraphics[width=0.5\linewidth]{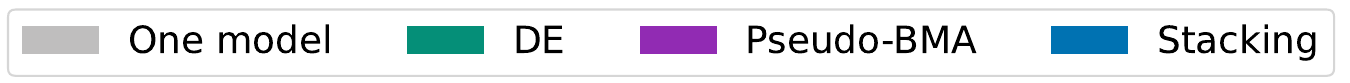}
    \caption{RMSE and CI obtained by a single mfVIR model, and by ensembling, stacking and pseudo-BMA when applied to mfVIR20 with Gaussian priors in the non-OOD (LHS) and 'complement-distributions' OOD (RHS) tasks. }
    \label{fig:metricsnasa}
\end{figure}

\textbf{General summary.} 
In the non-OOD scenario, stacking provides modest gains in both accuracy and uncertainty quantification, compared to deep ensembles and pseudo-BMA, with stable performance across different widths. \textbf{Instead, for the 'complement-distributions' OOD task, performance heavily depends on the network architecture.}  \textbf{Deep ensembles show the most stability, with stacking and pseudo-BMA becoming competitive as the width increases.}

\section{Discussion}
The message of an optimist's conclusion could question the common belief that the mean-field variational approximations are generally overly restrictive and do not capture the true posterior and the uncertainty well. Even with increases in computing power, the computational costs of sampling algorithms suggest that it may not be feasible for most modern neural networks and datasets. Moreover, although HMC is often considered as a gold standard, we have seen this may not be the case for BNNs due to complexity and multimodality of the posterior. 
Indeed, in a variety of experiments considered in \cref{sec:limitwidth,sec:limitdepth,sec:ood,sec:stackingexample} \textbf{mfVI overall provided better uncertainty quantification than HMC,} and in out-of-distribution settings, the empirical coverage of the latter was close to zero. We note that \textbf{for single-layer neural networks, HMC outperformed mfVI only in terms of accuracy.}  At the same time, for deeper networks and in out-of-distribution scenarios, the accuracy of mfVI was often comparable to HMC. Further, in \cref{sec:limitdepth} we confirmed that even for slightly deeper networks the time needed for HMC becomes a burden, which makes variational inference a very attractive alternative to sampling.  Nevertheless, in \cref{sec:limitwidth} we observed that \textbf{the restrictions imposed by the factorized families can obstruct models from effectively learning from the data}. In real-life scenarios where one is required to evaluate the future predictive performance of the model before applying it to the unseen data, the estimate of the expected log pointwise predictive density can serve as a reliable diagnostic and thus, PSIS-LOO estimates can be beneficial for model assessment and combination. In \cref{sec:stackingexample,sec:NASA}, \textbf{stacking and ensembles of BNNs were shown to be a possible solution when dealing with multimodal posteriors, helping to both improve accuracy and uncertainty quantification even in the extreme OOD scenario.}  We find that stacked or ensembled variational approximations are competitive to HMC at a much-reduced cost. Not only do the performance of individual models depend on architectural choices, but as observed in \cref{sec:NASA}, the model averaging techniques are themselves influenced by these modelling choices. Finally, we note that overall in our experiments, there was no considerable and systematic difference in the performance between the BNNs with Gaussian and Student-t priors.

This work highlights the model's sensitivity to architectural choices, namely, width, depth and activation function. Future work could study the performance of various more elaborate than Gaussian or Student-t choices of priors placed on the weights, including sparsity-inducing priors which have been shown to improve the accuracy and calibration \citep{blundell2015weight,polson2018posterior}. 
Further, an important avenue for research is to consider the so-called structured variational inference with less restrictive variational families, and more generally, study the trade-off between the expressiveness of the variational family and scalability. Finally, given the multimodal nature of distributions arising in Bayesian neural networks, a promising avenue for research is to continue improving model combination techniques. This includes developing a better understanding of the number of models required for optimal performance with existing ensembling methods, as well as exploring more advanced approaches such as adaptive variational Bayes frameworks \citep{ohn2024adaptive} or hierarchical stacking and pointwise model combination \citep{yao2022stacking}.

%

\bibliographystyle{plainnat}
\bibliography{refs}

\appendix
\section{Metrics and Practicalities}  \label{appendix:metrics}
Recall that we denoted the training data to be $\Data = \{x_n, y_n \}_{i=1}^N$ and the new data for testing to be $\tilde{\Data} = \{\tilde{x}_n,\tilde{y}_n,\}_{n=1}^{\tilde{N}}$.
Denote the approximated posterior predictive mean as $\by$, the set of $S$ samples of the signal as $\bmu^{S}$ and of the observations as $\by^{S}$.  Upon computing the posterior predictive distribution, we obtain the root mean squared error and the empirical coverage as: 
\begin{align*}
    \RMSE & = \sqrt{\frac{1}{N} \sum_{n}^N \left [(\tilde{y}_n - \E^{S}[y^{S}_{i}])^2\right ]},\\
    \text{EC}& =\frac{ \# \{ \by_ \in [q_{0.025}, q_{0.975}]\}}{N}, \text{ where } q \text{ are quantiles of } \bmu^S \text{ or } \by^S.
\end{align*}
The results are obtained when the number of iterations of mfVI is set $10^4$, $10^4$, $5\times10^4$, $6\times10^4$ to train models with, respectively, $20, 200, 1000, 2000$ hidden units in a layer. In models trained with HMC, the number of samples used for warmup was set to $10^3$, the samples used for posterior is $10^3$ in models with $20$ hidden units, and $2\times10^3$ in models with $200, 1000, 2000$ hidden units in a layer. Across all the experiments with simulated data, the learning rate of the optimizer is set to $5e-3$.
In deeper networks, the number of iterations of mfVI was set to $L \times 10^4$; the HMC had the number of warmup samples fixed to $10^3$ and the number of samples was $\min (4\times 10^3, L \times 10^3)$.

When evaluating BNN's performance on the NASA's rocket booster simulation, we only considered mfVI, and set the number of iterations to $10^4$ and learning rate to $5e-2$, with one exception of the OOD scenario and 200 hidden units, when we trained the model for $2\times10^4$ iterations.

\textbf{Remark on the initialization.} Based on the empirical evidence, we observed that in our experiment for $L = 1, 2$ the NumPyro implementation of mfVI requires the initialization mode to be set to "init to feasible", which chooses the initialization point uniformly (ignoring the prior distribution). Whereas for $L =3,4,5,6$ mfVI requires "init to mean", which sets initial parameters to the prior mean, and "init to feasible" will fail. Conversely, the NumPyro implementation of the HMC fails if the initialization location is set to "init to mean" but performs fine if it is always set to "init to feasible", i.e. ignoring the distribution parameters.

\section{Supplementary to Predictive Model Assessment}  \label{appendix:predictivemodelassesment}
\subsection{Empirical model assessment}
\begin{figure}[t!]
    \centering
    \subcaptionbox{Posterior predictive checks for the wider models based on the kernel density estimates of the observed $\by$. \label{oodppc2000}}[.4\textwidth]{ \includegraphics[width=0.9\linewidth]{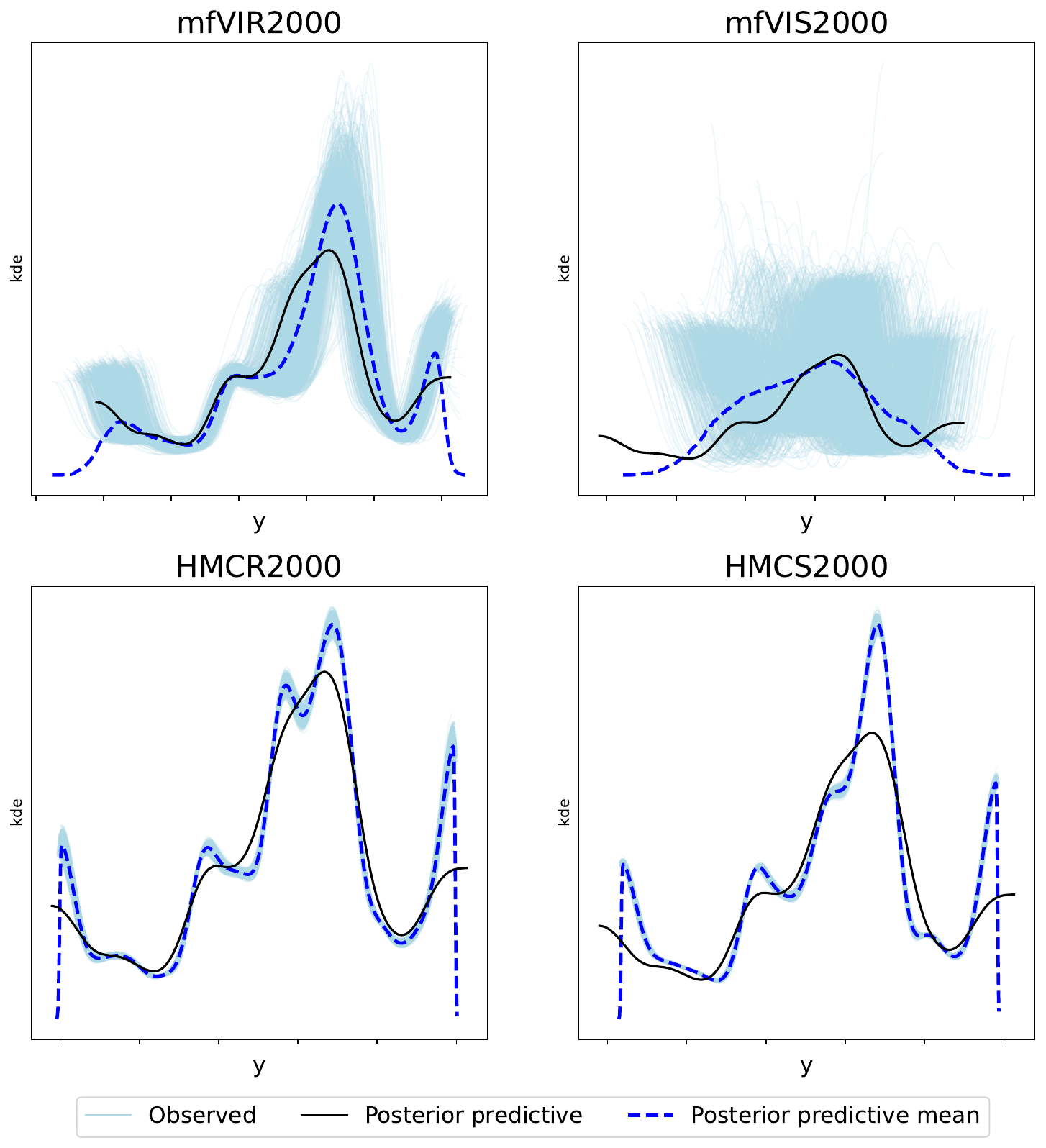}}
    \subcaptionbox{Posterior predictive distribution and posterior predictive mean compared to the observed $\by$. \label{oodppc2000lm}}[.4\textwidth]{ \includegraphics[width=0.9\linewidth]{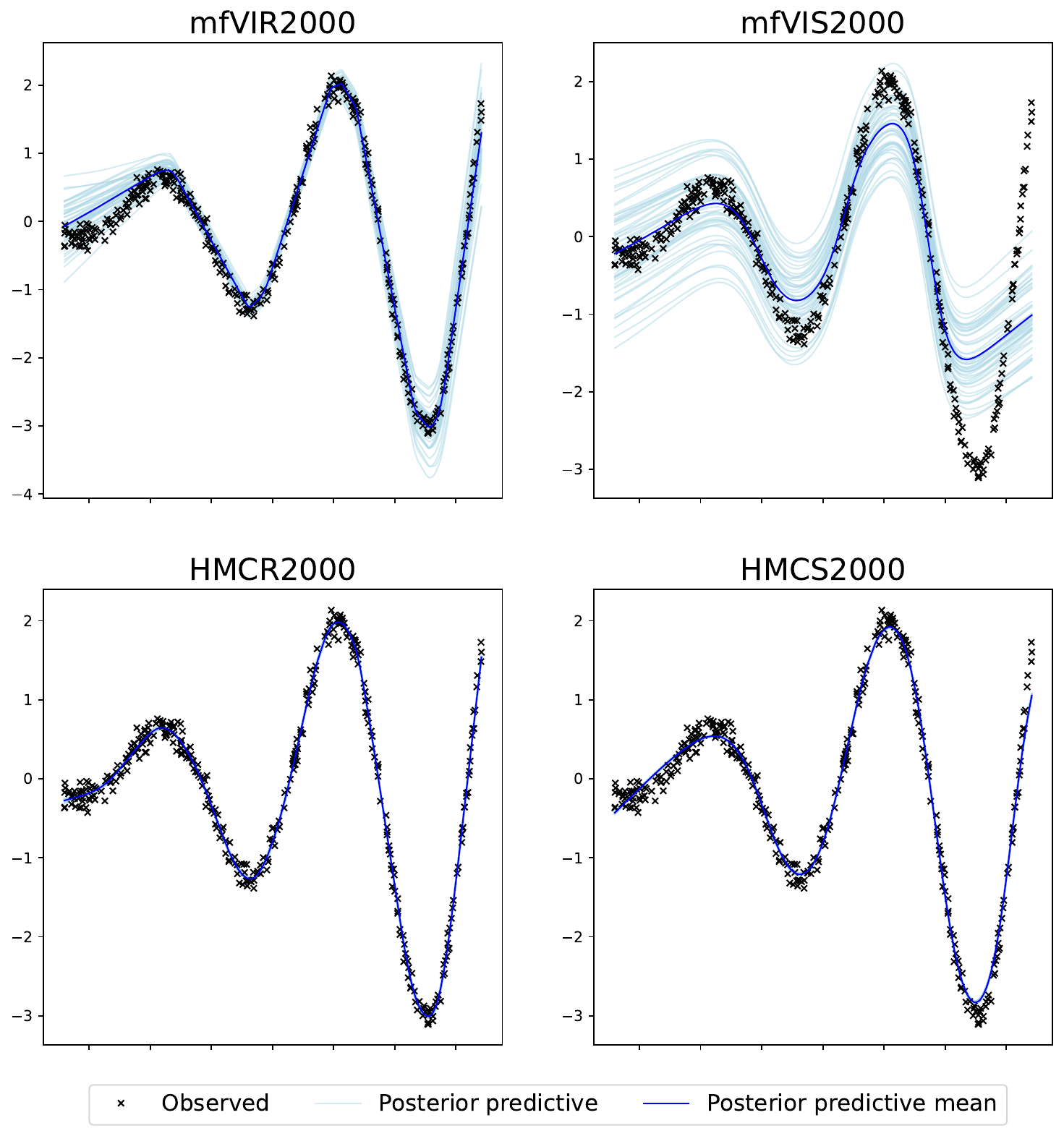}}
    \subcaptionbox{The correspondence between the $\loo$ and the $\RMSE$ in the OOD scenario. Higher $\loo$ should correspond to lower $\RMSE$.  \label{elpdvsrmse}}[.8\textwidth]{ \includegraphics[width=\linewidth]{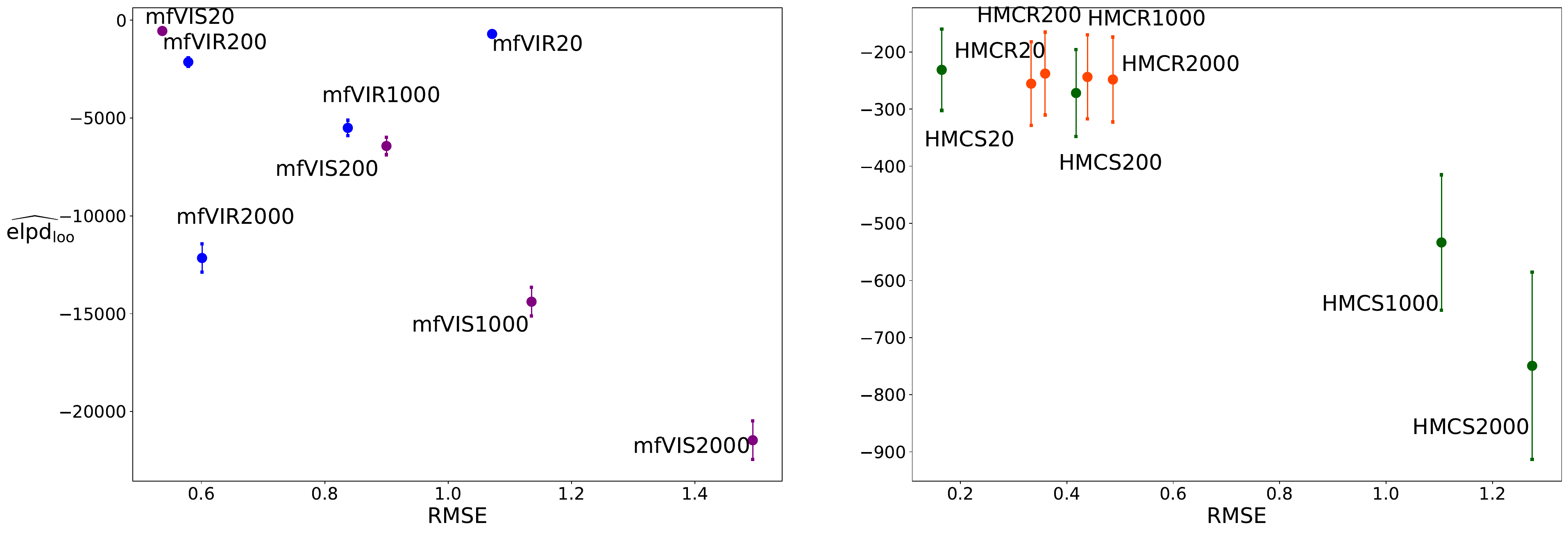}}
    \caption{Estimating the out-of-distribution performance of models with Gaussian priors before seeing the new data: testing the (a), (b) PPC  and (c) $\loo$. The mfVIR2000 is confirmed to be unreliable in all methods. The PPC of the HMCS2000 does not provide enough information to judge its performance in the OOD settings, while the $\loo$ does.}
    \label{ppcandelpd}
\end{figure}

\begin{figure}[ht!]
    \centering
    \subcaptionbox{Posterior predictive checks for the wider models based on the kernel density estimates of the observed $\by$. \label{oodppc2000st}}[.4\textwidth]{ \includegraphics[width=0.9\linewidth]{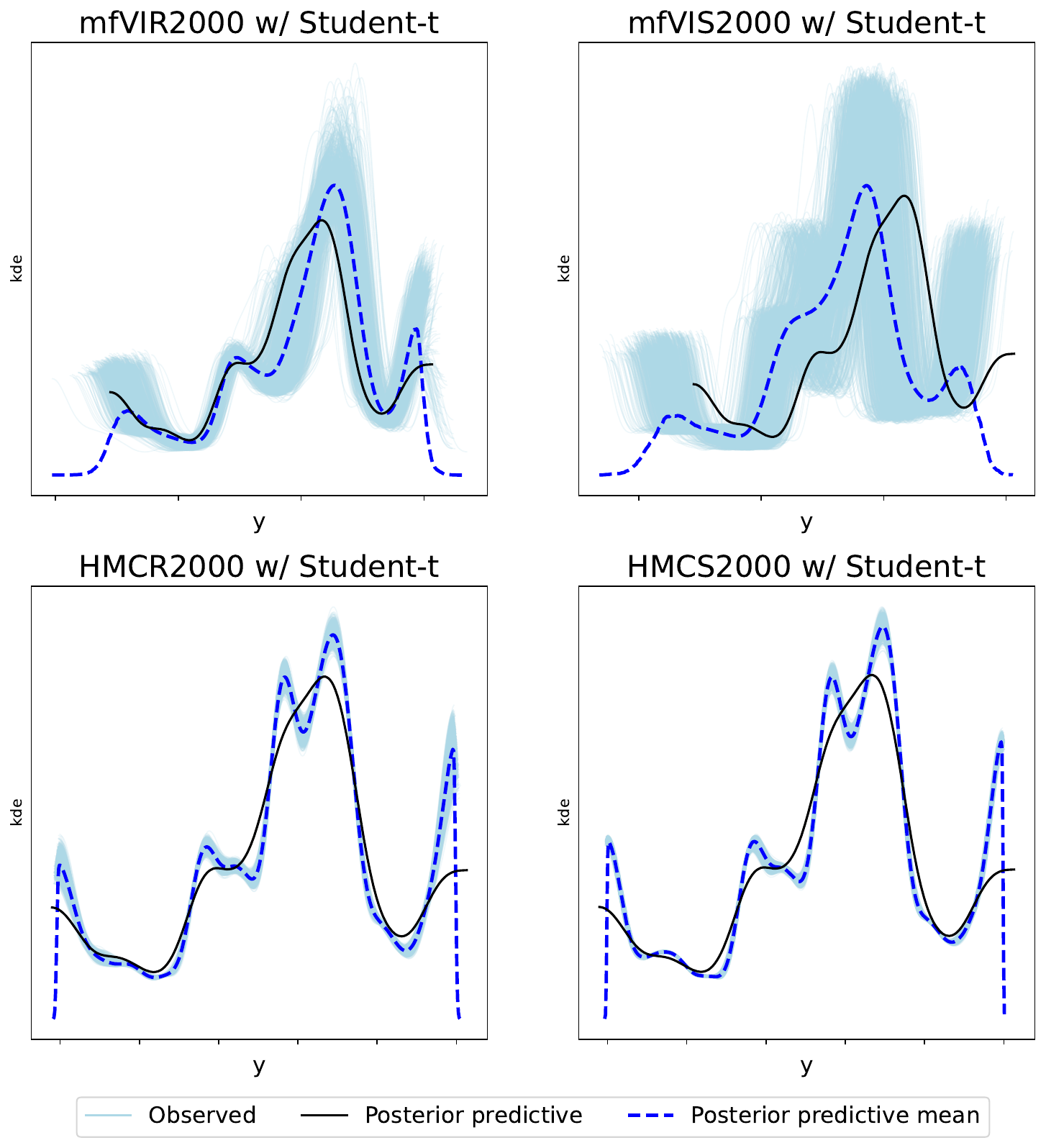}}
    \subcaptionbox{Posterior predictive distribution and posterior predictive mean compared to the observed $\by$. \label{oodppc2000lmst}}[.4\textwidth]{ \includegraphics[width=0.9\linewidth]{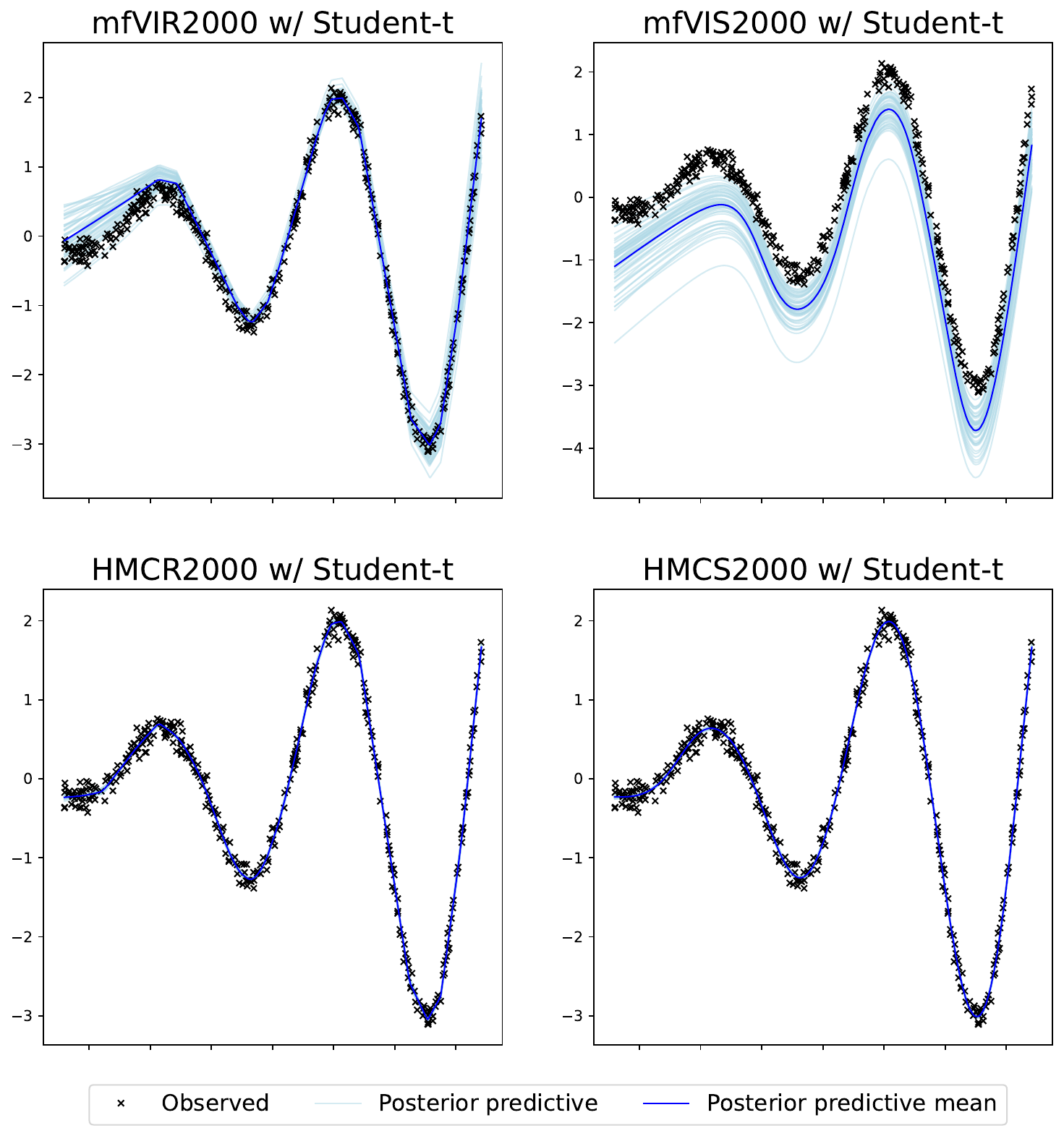}}
    \subcaptionbox{The correspondence between the $\loo$ and the $\RMSE$ in the OOD scenario. Higher $\loo$ should correspond to lower $\RMSE$.  \label{elpdvsrmsest}}[.8\textwidth]{ \includegraphics[width=\linewidth]{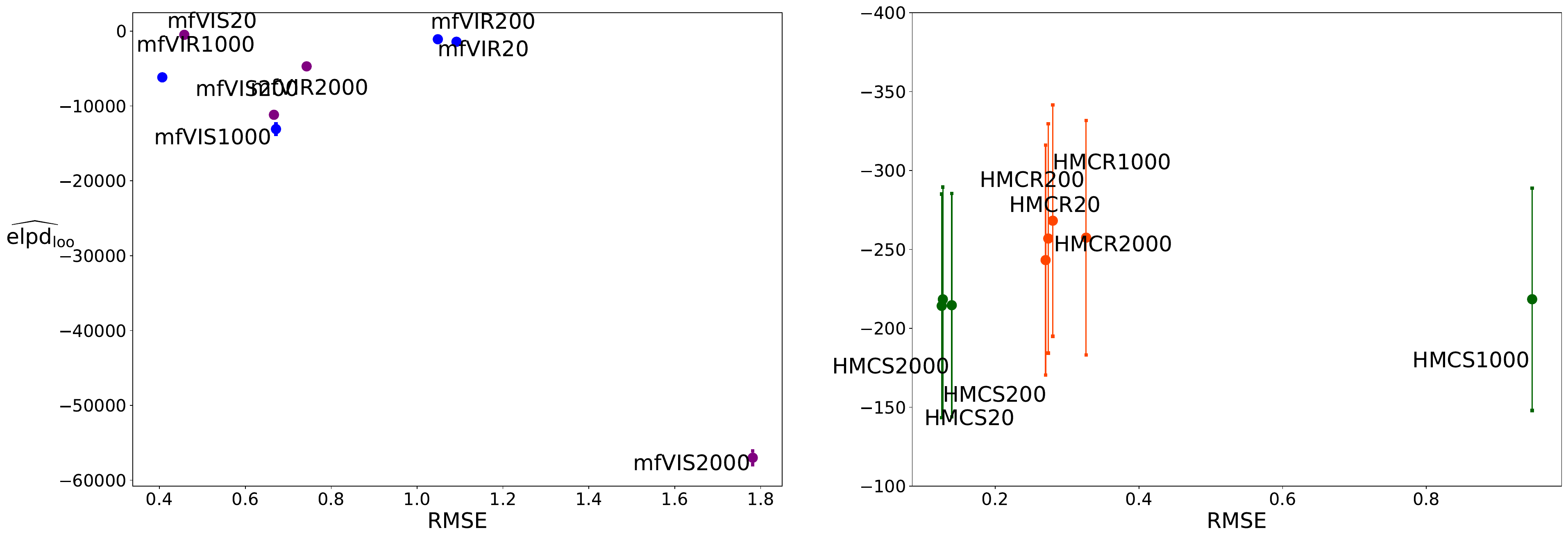}}
    \caption{Estimating the out-of-distribution performance of models with Student-t priors before seeing the new data: testing the (a), (b) PPC  and (c) $\loo$. The mfVIR2000 is confirmed to be unreliable in all methods. The PPC of the HMCS2000 does not provide enough information to judge its performance in the OOD settings, while the $\loo$ does.}
    \label{ppcandelpdst}
\end{figure}

Recall, the OOD scenario, where we have specifically created an extreme case when the training data $\Data_c$ and the new data $\tilde{\Data}_c$ were covering disjoint regions of the true function. 
To asses the expected out-of-sample predictive performance, we could begin with posterior predictive checks (PPC), which compare the true $\Data_c$ to datasets simulated from the posterior predictive distribution \citep{gelman2020bayesianworkflow}. \cref{oodppc2000} and \cref{oodppc2000st} provide the PPC based on the kernel density estimates of the observed $\by$, and \cref{oodppc2000lm} and \cref{oodppc2000lmst} compare posterior predictive to the observed $\by$ for all of the models with $D_1 = 2000$; both figures evidently classify the mean-field VI with sigmoid activation as inappropriate. However, it is not apparent that HMCS2000 is significantly inferior to HMCR2000. 
We compare the $\RMSE$ computed in the OOD settings to $\loo$ estimates.  Since the idea of estimating the expected log predictive density is in evaluating future predictive performance, we expect the higher $\loo$ to correspond to a better model and lower $\RMSE$. Indeed, \cref{elpdvsrmse} and \cref{elpdvsrmsest} are more informative in this sense than PPC diagnostics, and we observe the inverse dependence between $\loo$ and $\RMSE$. Sampling and approximation techniques result in different scales of $\loo$ estimates (in general, $\loo$ is lower for VI, especially, in wide networks) and thus, we compare the models trained with different algorithms for better visualization.

\textbf{General summary.} 
In certain cases, such as mfVIS for large width, the posterior predictive checks are able to detect an undesirable model. However, when the PPCs are not sufficient, we confirmed that the PSIS-LOO estimates of the expected log pointwise predictive density can serve as robust diagnostics for both the mfVI and the HMC methods.

\subsection{Correspondence Between WAIC and RMSE}\label{app:architectureexperimentWAICRMSE}
\begin{figure}[ht!]
    \centering
    \includegraphics[width=0.45\linewidth]{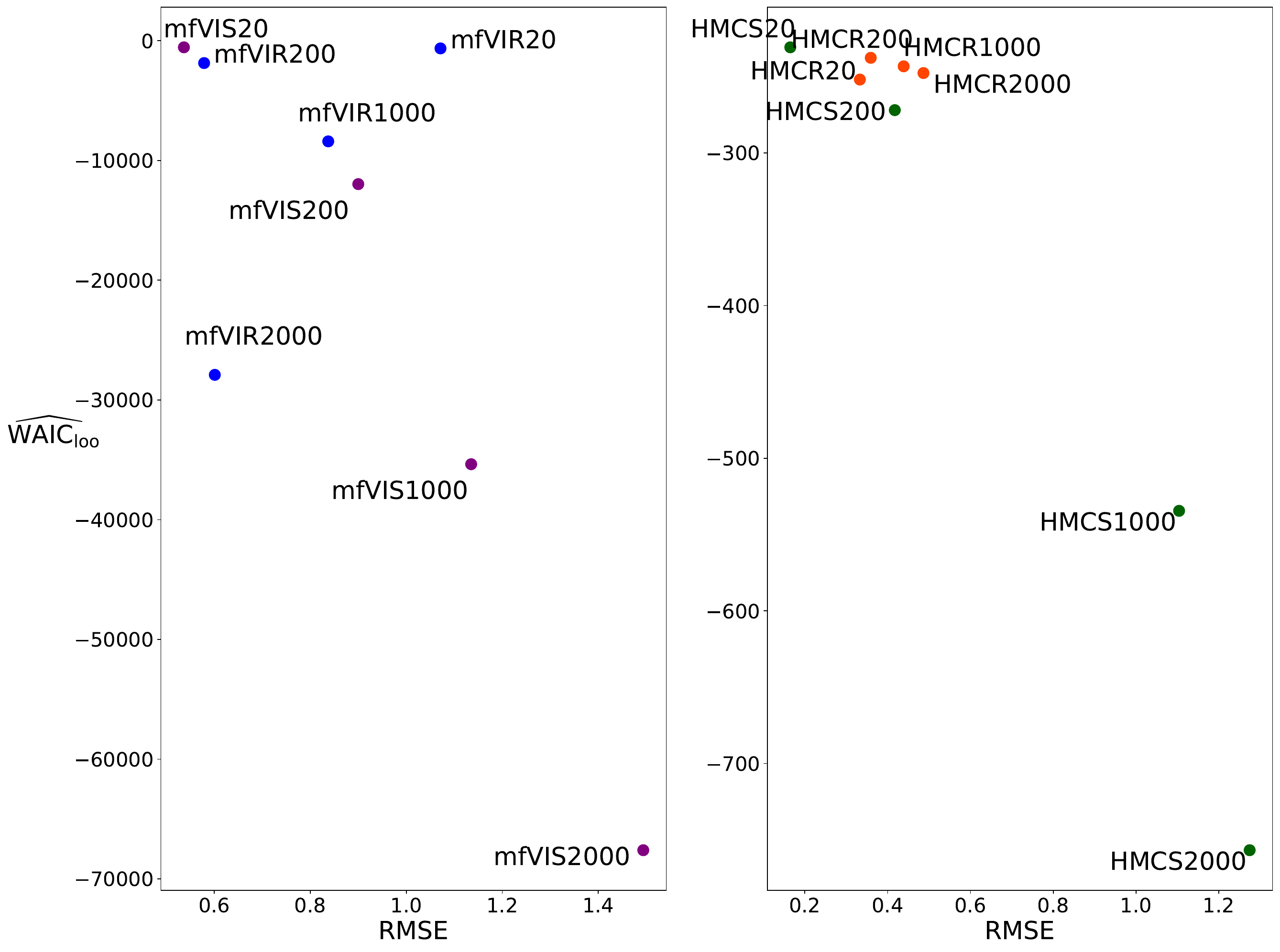}
    \caption{Estimating the out-of-distribution performance before seeing the new data: the correspondence between the $\widehat{\text{elpd}}_{\text{WAIC}}$ and the $\RMSE$ in the OOD scenario. Similarly to $\loo$, the higher $\widehat{\text{elpd}}_{\text{WAIC}}$ should correspond to lower $\RMSE$. }
    \label{fig:waic}
\end{figure}
Above, we compared the RMSE obtained in the OOD experiment to the estimates of the expected log pointwise predictive density obtained with LOO-CV. An alternative approach is to overestimate the elpd by first computing the log pointwise predictive density of $\Data$ and then adjusting it by some correction term.
Specifically, suppose that $\btheta$ are the parameters of the model and $\btheta^s, s = 1, \ldots, S$ are simulation draws. Then, one can evaluate the log pointwise predictive density as
\begin{align*}
    \widehat{\text{lpd}} &= \sum_{n=1}^N \log \left [\frac{1}{S} \sum_{s=1}^S p (y_n | \btheta^s)\right ].
\end{align*}
With $\widehat{\text{lpd}}$ at hand, a superior successor of the Akaike Information criterion (AIC) \citep{AIKAkaike1998} and the Deviance information criterion (DIC) \citep{DICspiegelhalter2002bayesian} called the  Watanabe-Akaike information criterion (WAIC) \footnote{Also called Widely Applicable information criterion} \citep{watanabe10a} can be obtained. To mitigate the bias between the log pointwise density and the expected utility, WAIC subtracts the simulation-estimated effective number of parameters:
\begin{align*}
    \widehat{\text{elpd}}_{\text{WAIC}} = \widehat{\text{lpd}} - \widehat{p}_{\text{WAIC}}, \text{ where } \widehat{p}_{\text{WAIC}} = \sum_{n=1}^N \Var^S(p (y_n | \btheta^s)). 
\end{align*}

Here, $\Var^S$ is the sample variance and the estimated effective number of parameters $\widehat{p}_{\text{WAIC}}$ can be seen as a measure of model complexity. Asymptotically, WAIC is equivalent to the Bayesian leave-one-out cross-validation (LOO-CV) estimate of the expected utility \citep{watanabe10a}. Even though cross-validation is a natural framework for accessing the model's predictive performance, the WAIC was for a long time preferred over the LOO-CV due to the computational challenges arising from multiple model runs \citep{gelman2014understanding}. Moreover, whilst both the PSIS-LOO and WAIC estimates give nearly unbiased estimates of the predictive ability of the model, $\loo$ was shown to be more robust than $\widehat{\text{elpd}}_{\text{WAIC}}$; in the presence of limited sample size and weak priors, WAIC can severely underestimate $\hat{p}_{\text{WAIC}}$ and often has a larger bias towards the log predictive density \citep{VehtariGelmanGabry_2016,gelman2014understanding}. 

That said, computing $\loo$ involved Pareto smoothed importance sampling, and in some of the models, the estimated shape parameter of the generalized Pareto distribution gave a warning about the reliability of the LOO estimate. Here, we do an extra step and check if the WAIC and LOO estimates of the elpd agree in the case of models with Gaussian priors. \cref{fig:waic} illustrates the reverse dependency between the RMSE and $\widehat{\text{elpd}}_{\text{WAIC}}$ and is largely identical to the figure illustrating the dependency between the RMSE and $\loo$  (in terms of the location of coordinates but not the error bars). Therefore, we can conclude that LOO estimates of the elphd can be seen as relatively reliable.

\section{Supplementary for Student-t Priors}
\label{appendix:studentt}
\begin{figure}[ht!]
    \centering
    \includegraphics[width = 0.4\linewidth]{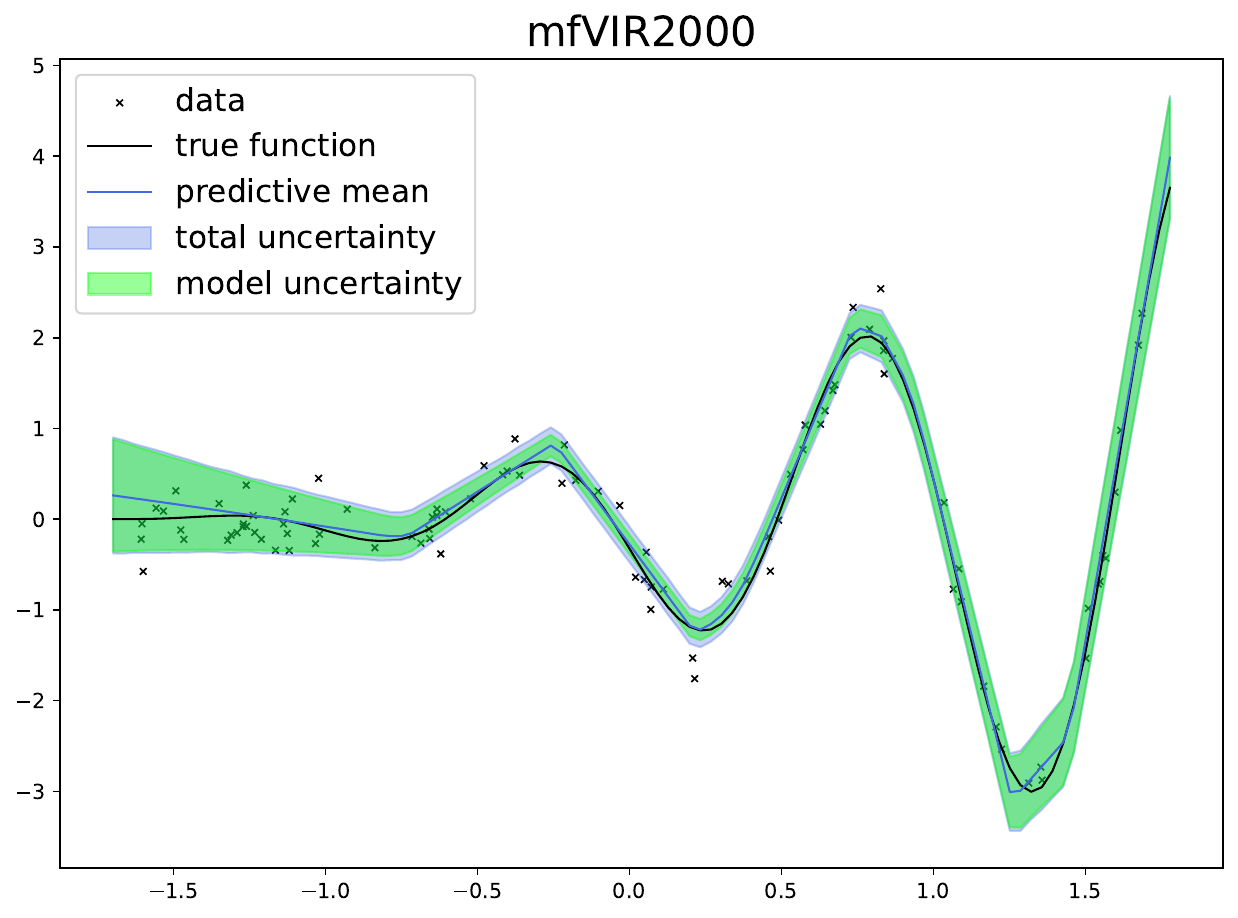}
    \includegraphics[width = 0.4\linewidth]{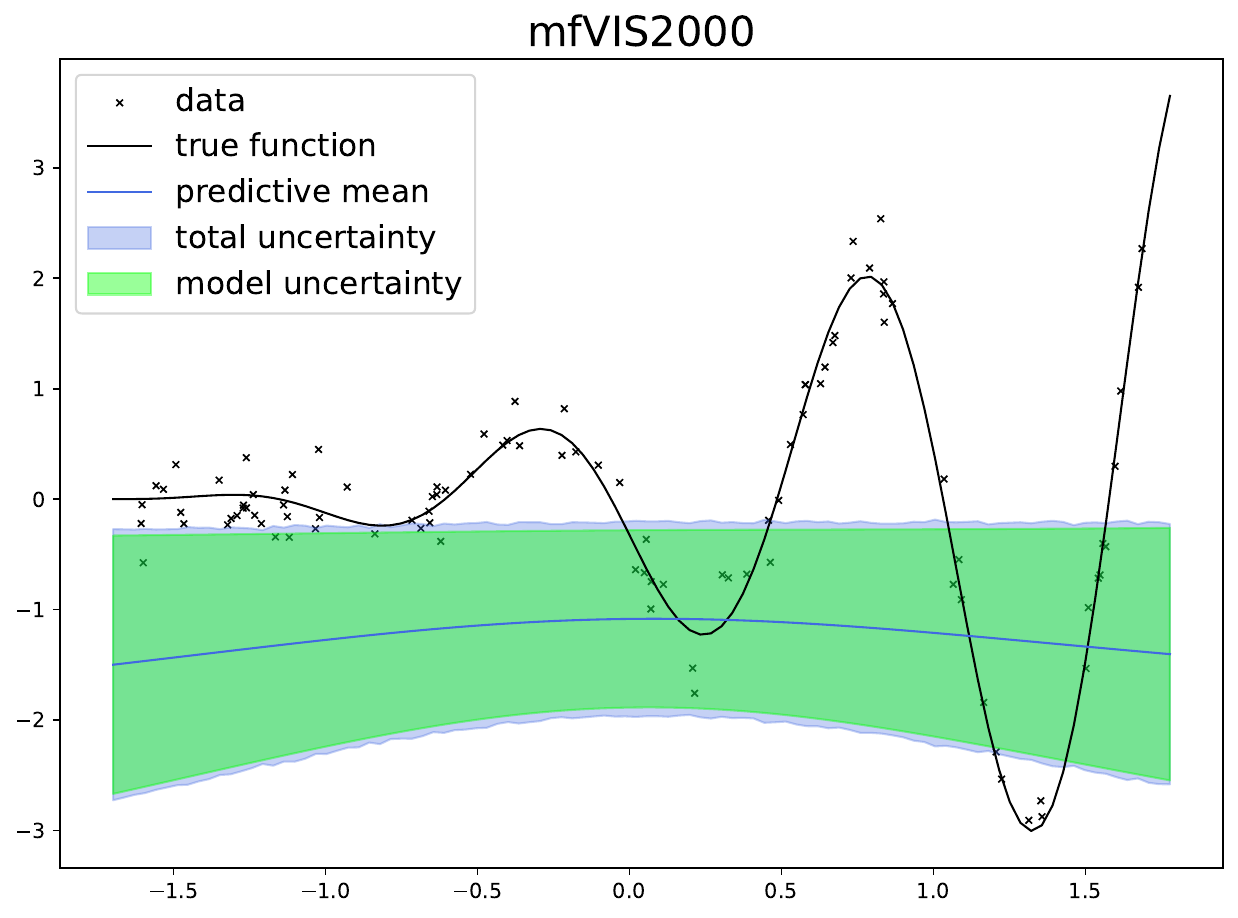} \\
        \includegraphics[width = 0.4\linewidth]{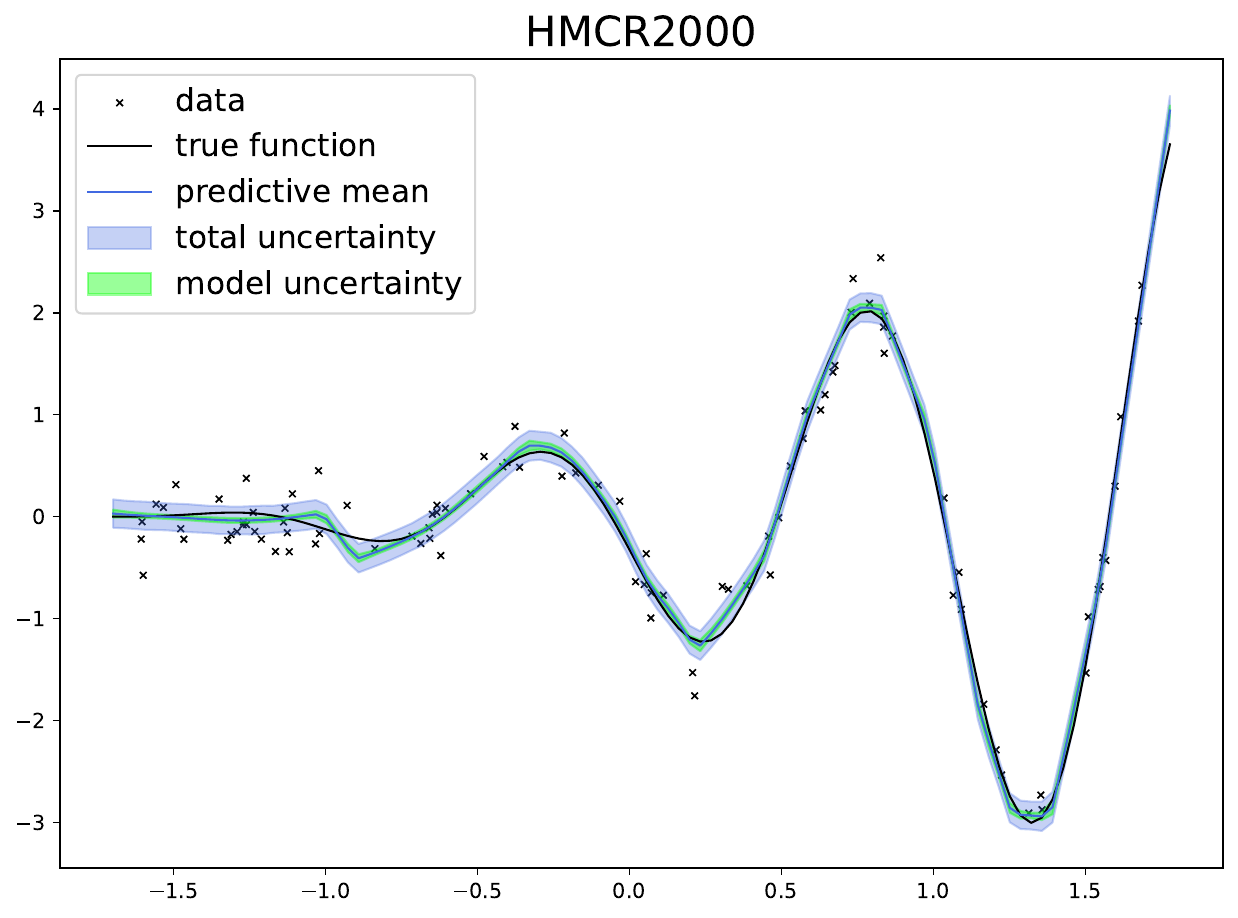}
   \includegraphics[width = 0.4\linewidth]{supplement_figures/stHMCR2000.pdf}
\caption{ The predictions and uncertainty estimates obtained by each model for $D_1=2000$ and Student-t priors. }
\label{fig:2000plotsst}
\end{figure}

In the main body of the work, we first studied the performances of mfVIR, mfVIS, HMCR and HMCS with 1 hidden layer and either Gaussian or Student-t priors as the width increases. The predictions of all four models with Student-t priors when $D_1=2000$ are provided on the \cref{fig:2000plotsst}. Similar to models with Gaussian priors, the performance of the mfVIS dips with the increase in the dimension of the hidden layer; for wider networks, its posterior predictive distribution fails to capture the data, and degenerates to the prior. The HMCR2000 model exhibits a better performance than the HMCS2000 both in terms of accuracy and uncertainty quantification.  

\begin{figure}[ht!]
    \centering
    \includegraphics[width = 0.4\linewidth]{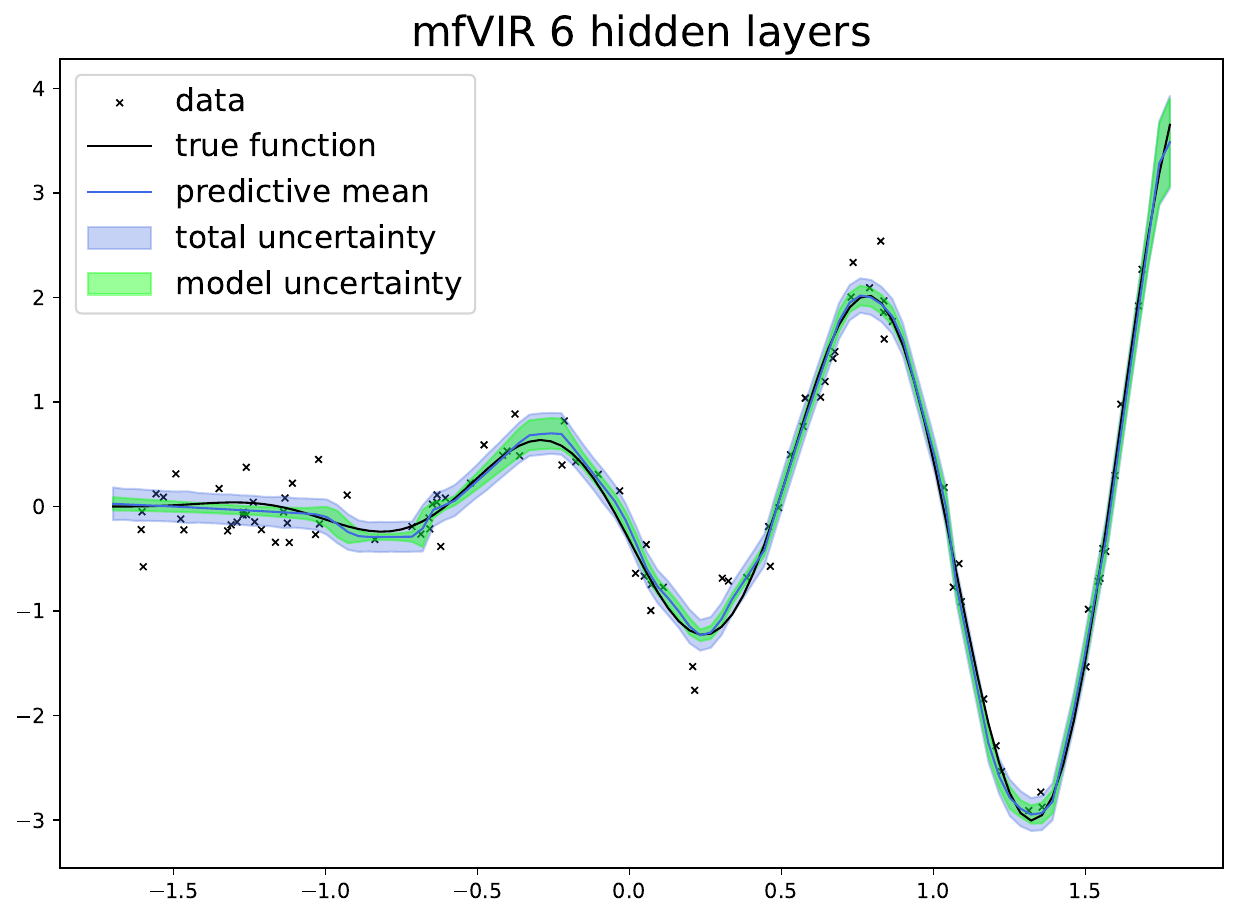}
    \includegraphics[width = 0.4\linewidth]{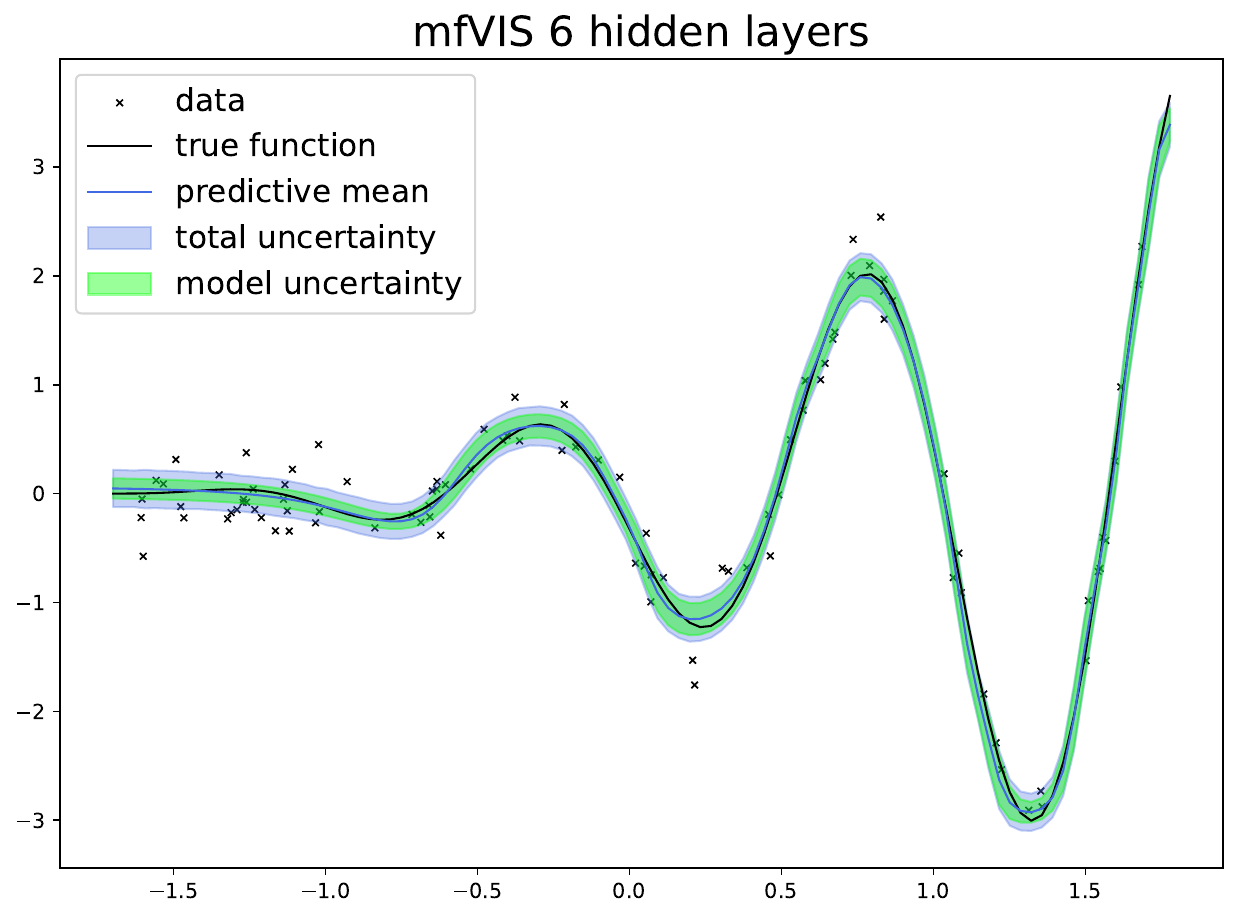} \\
        \includegraphics[width = 0.4\linewidth]{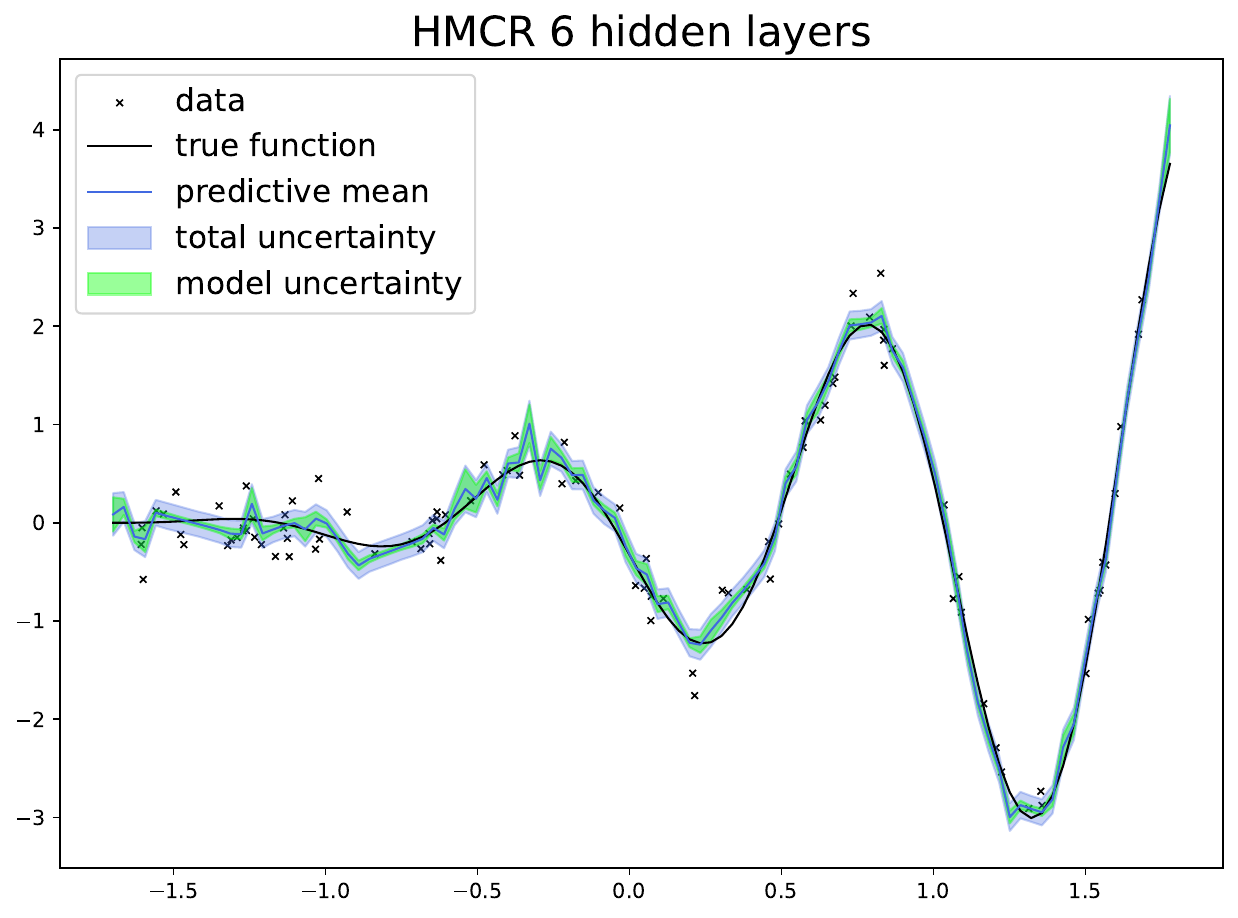}
   \includegraphics[width = 0.4\linewidth]{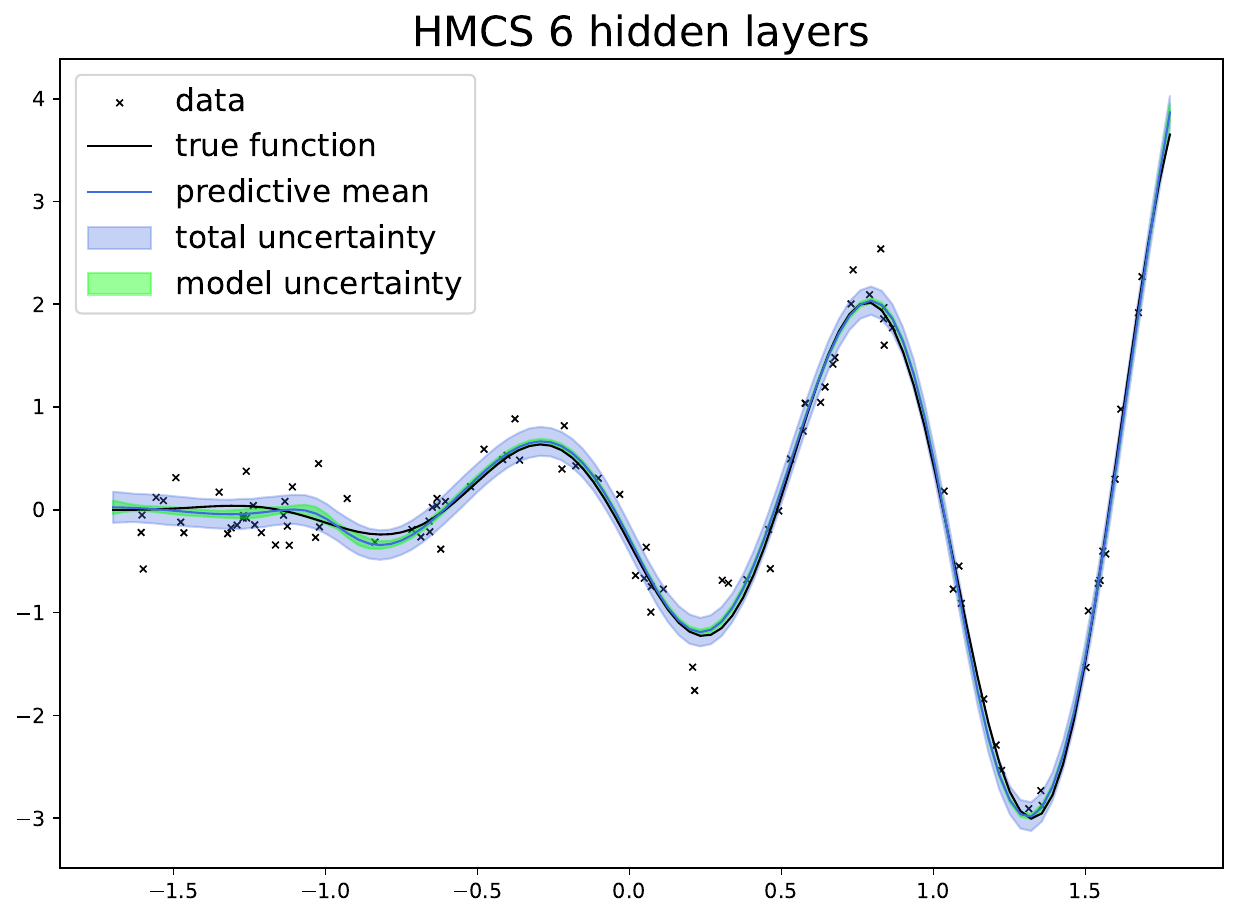}
\caption{ The predictions and uncertainty estimates obtained by each model for $L=6$ and Student-t priors. }
\label{fig:stdepthex}
\end{figure}

Consider BNNs with Student-t priors with the number of layers $L$ varying from 1 to 6 and a fixed number of hidden units in each layer $D_h =20$, then \cref{fig:stdepthex} illustrates the predictions of all four models with $L=6$

\begin{figure}[ht!]
    \centering
    \includegraphics[width = 0.4\linewidth]{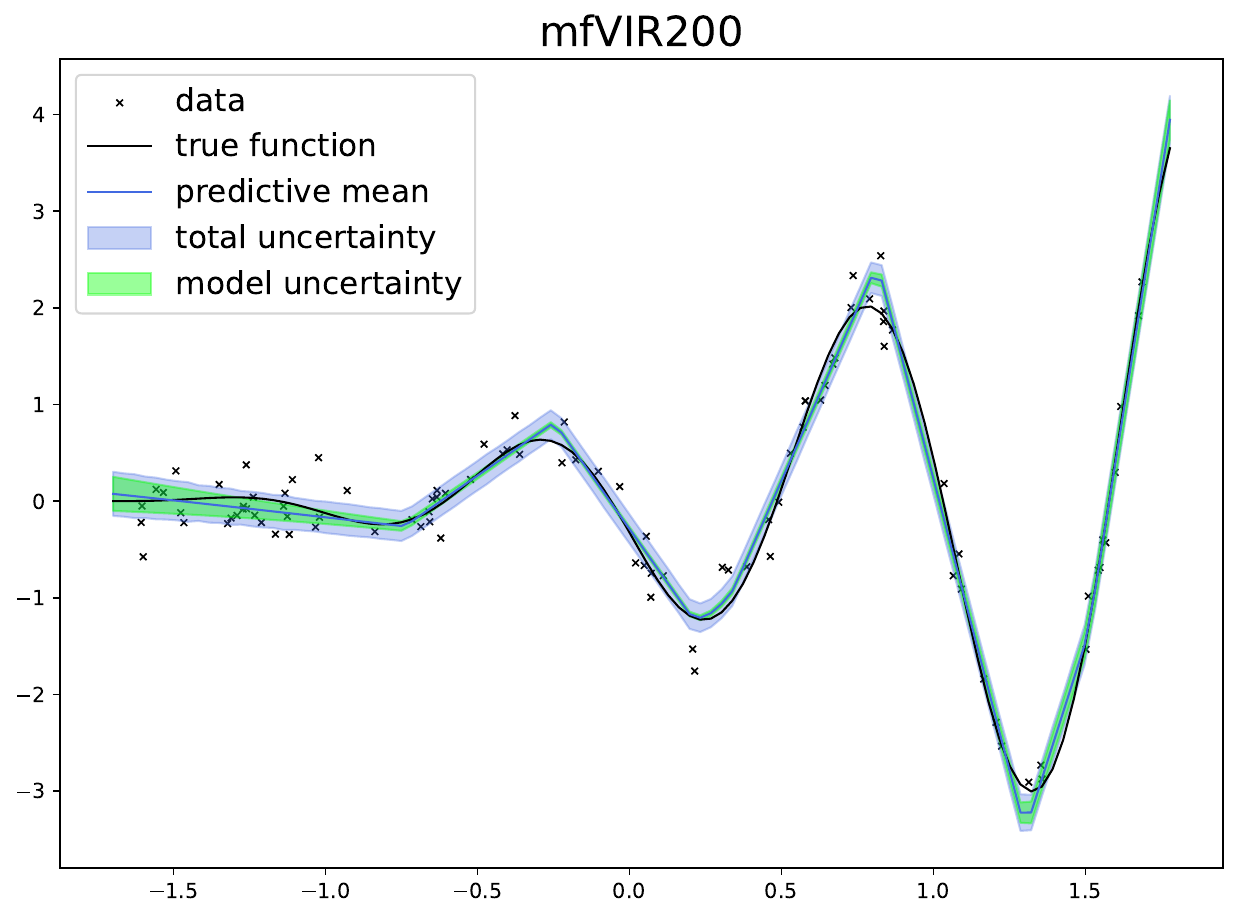}
    \includegraphics[width = 0.4\linewidth]{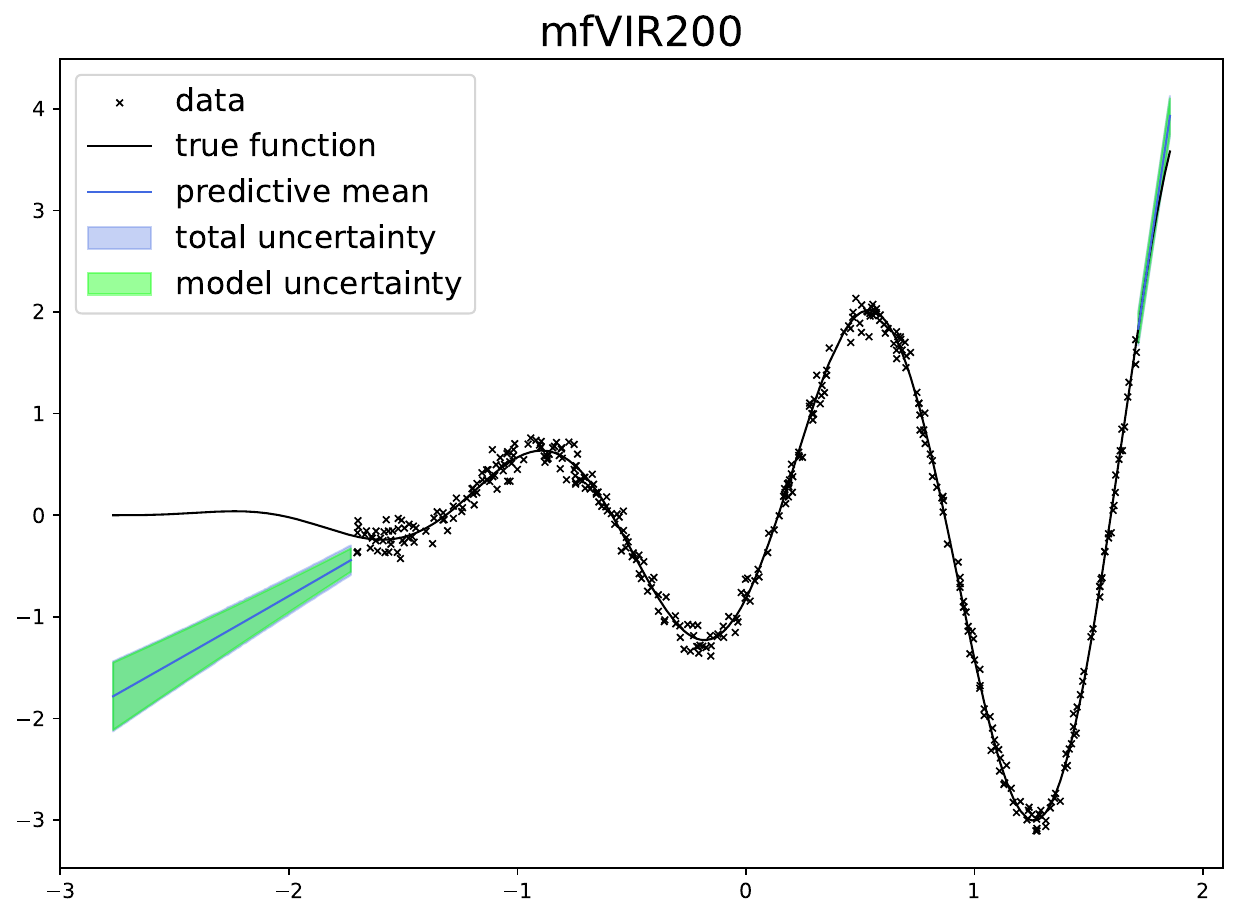}
     \\
        \includegraphics[width = 0.4\linewidth]{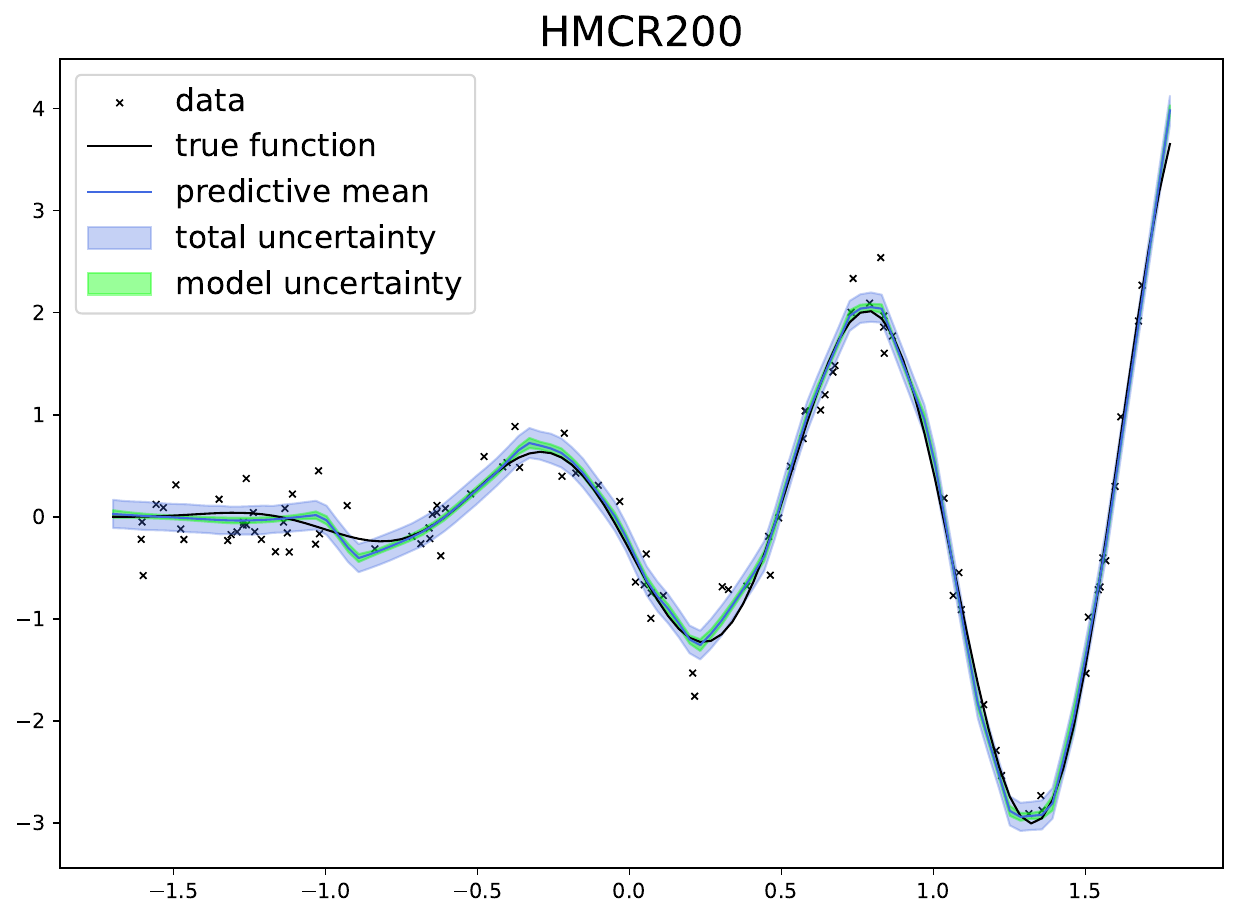}
   \includegraphics[width = 0.4\linewidth]{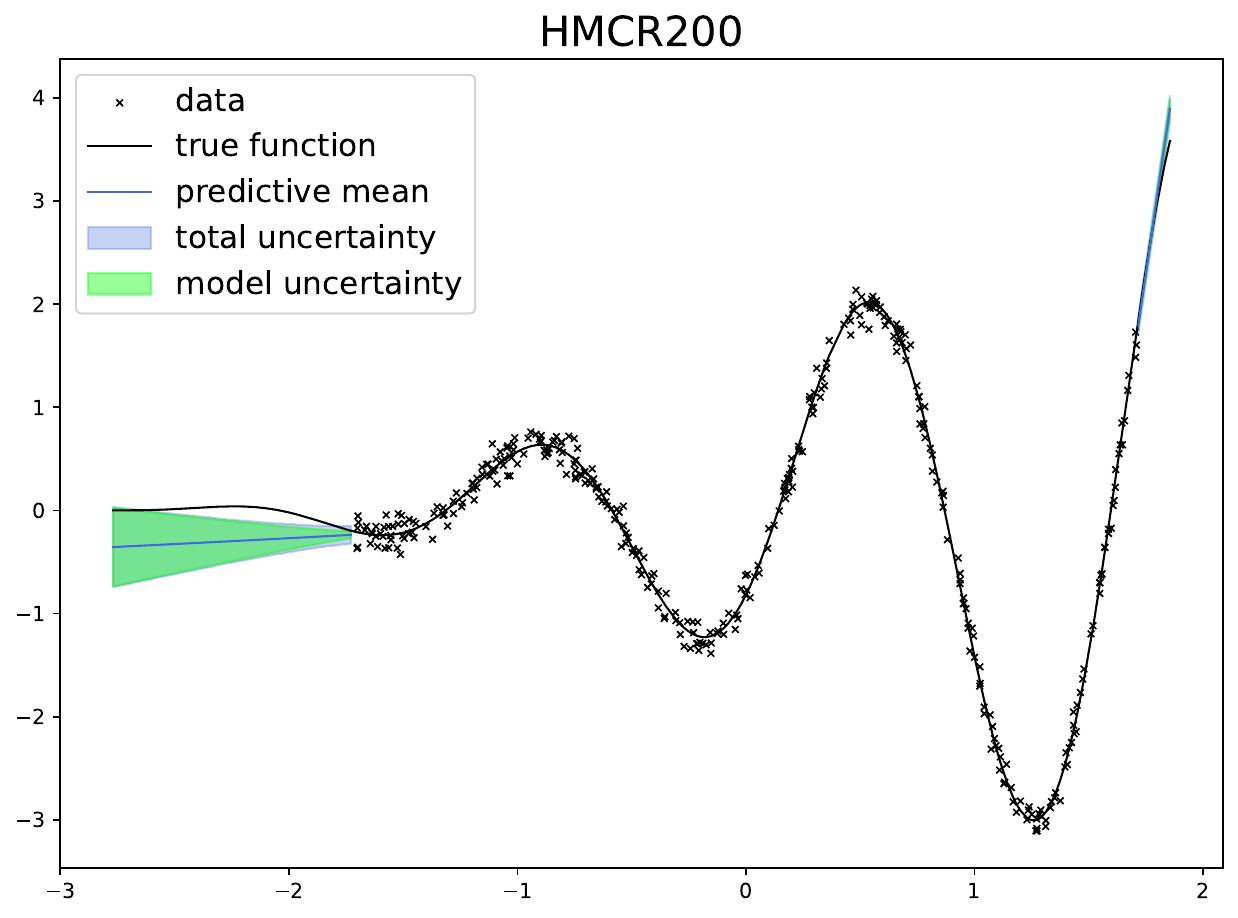}
\caption{  Within-the sample and out-of-sample predictions and uncertainty estimates  of BNNs with $D_1=200$, ReLU activation and Student-t priors. }
\label{fig:oodst}
\end{figure}

Further, \cref{fig:oodst} compares non-OOD and OOD predictions obtained by the BNNs with ReLU activation, Student-t priors and $D_1=200$.

\section{Supplementary to Ensembles and Averages} \label{appendix:ensemblesandaverages}
\textbf{Remark on constructing deep ensembles.}
Given $\mathcal{M} = \{M_1, \ldots, M_K\}$ a collection of models suppose that $K$ approximations $\tilde{\by}_k$ of the posterior $p(\tilde{\by} | \Data, M_k)$ have means $\mu_k$ and variances $\sigma_k^2$ or $k=1, \ldots, K$. Then the mean and variance of an ensemble of approximations:
\begin{align*}
    \mu_{\text{DE}} = K^{-1}\sum_{1}^K\mu_k, \quad
    \sigma_{\text{DE}}^2 = K^{-1}\sum_{1}^K\sigma_k^2 + \mu_k^2 - \mu_{\text{DE}}^2.
\end{align*}
In general, given weights $\omega_k = p(M = M_k)$ the mean and the variance are 
\begin{align*}
     \mu_{\text{DE}}& = \E [\E [\tilde{\by} | M]]=\sum_{1}^K\omega_k\mu_k, \\
    \sigma_{\text{DE}}^2 &= \E[\E[\tilde{\by}^2| M]] - \mu_{\text{DE}}^2 = \sum_{1}^K\omega_k\E[\tilde{\by_k}^2] - \mu_{\text{DE}}^2\\
    & = \sum_{1}^K\omega_k(\sigma_k^2 + \mu_k^2) - \mu_{\text{DE}}^2.
\end{align*}
\subsection{Models with Student-t priors}
\begin{figure}[t]
    \centering
     \includegraphics[clip, trim=0.0cm 2cm 0.0cm 0.0cm, width=0.9\linewidth]{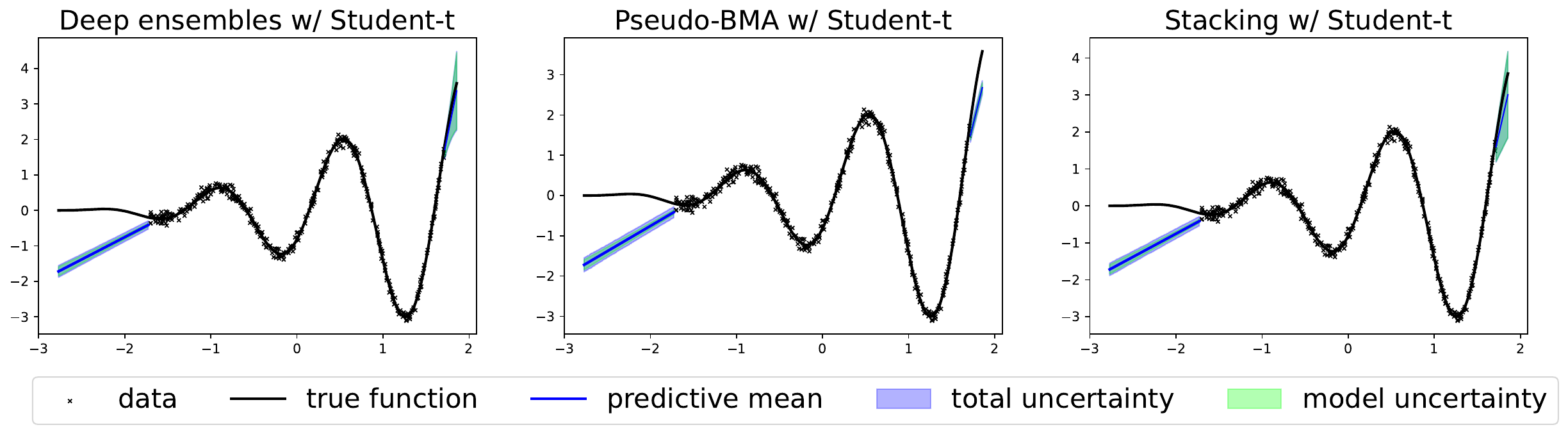}
       \includegraphics[width=0.9\linewidth]{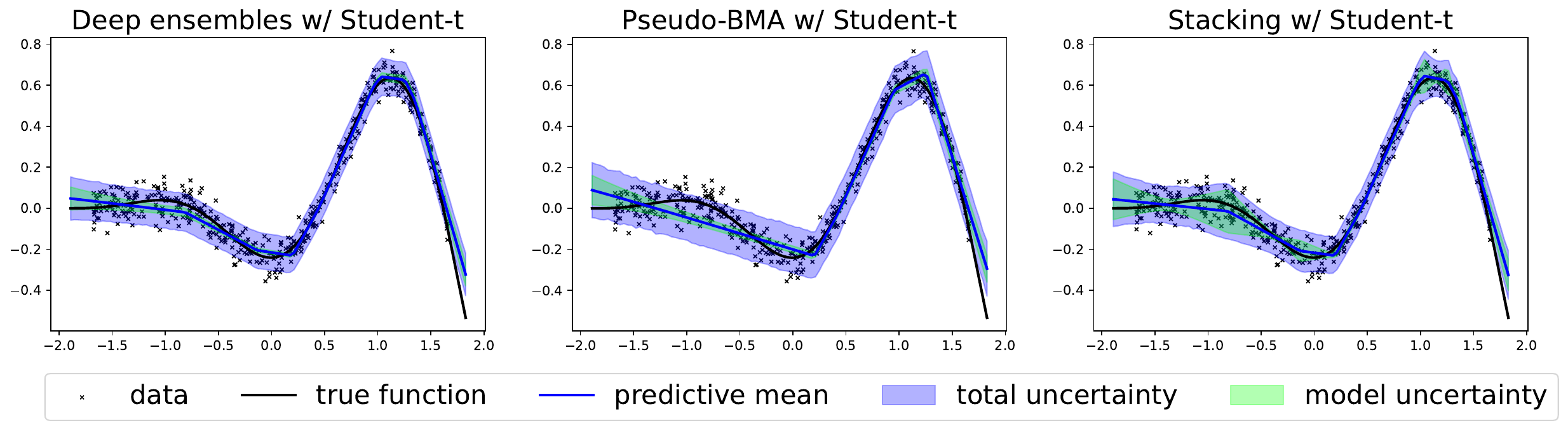}
    \caption{Predictions obtained by ensembling, stacking and pseudo-BMA when applied to mfVIR20 with Student-t priors in the 'complement-distributions' (top)  and 'related-distributions' (bottom) OOD tasks. The pseudo-BMA is worse than DE and stacking, which are very similar to each other.}
    \label{fig:OODstackst}
\end{figure}
Consider the mfVIR20 model with Student-t priors, the 'complement-distributions' and the 'related-distributions' OOD tasks. In each task, we obtain 10 posterior predictive distributions and construct ensemble, pseudo-BMA and stacking approximations, the results are illustrated \cref{fig:OODstackst}.

\subsection{Deeper Networks} \label{appendix:deepernetworks}
\begin{figure}[ht!]
    \centering
    \includegraphics[clip, trim=0.0cm 2cm 0.0cm 0.0cm,width=0.9\linewidth]{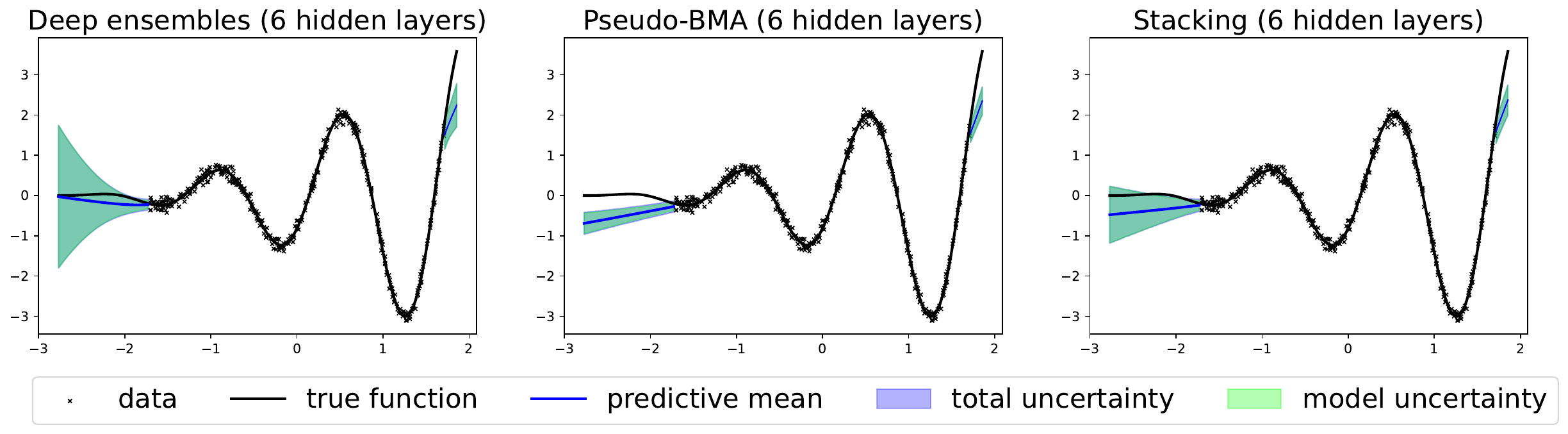}
    \includegraphics[width=0.9\linewidth]{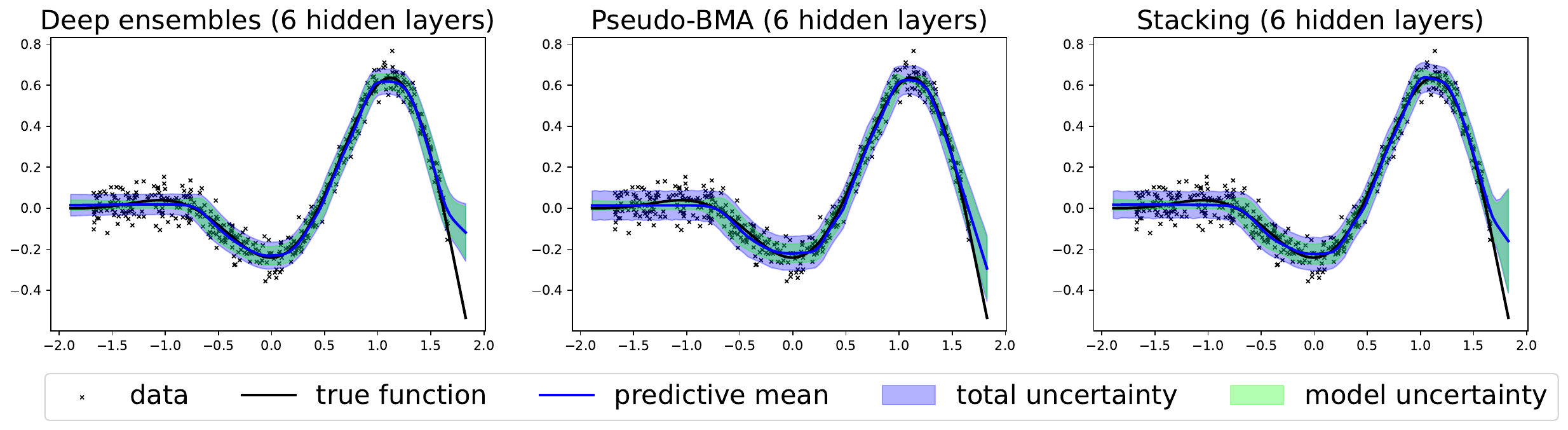}
    \caption{Predictions obtained by ensembling, stacking and pseudo-BMA when applied to mfVIR20 with $L=6$ in the 'complement-distributions' (top) and the 'related-distributions' (bottom) task.}
    \label{fig:stackmfvifor6layers}
\end{figure}
Based on the 10 posterior predictive distributions obtained starting from 10 different random initialization points, we construct ensemble, pseudo-BMA and stacking approximations for mfVIR and mfVIS models with Gaussian priors and $L=6$ hidden layers.  The results are consistent with the observation made in the main body of the work; in the 'complement-distributions' task (\cref{fig:stackmfvifor6layers}) pseudo-BMA is confirmed to be inferior to stacking and deep ensembles of BNNs. In the related distribution task (\cref{fig:stackmfvifor6layers}), in terms of both accuracy and uncertainty quantification, stacking is preferable over deep ensembles and pseudo-BMA with the latter performing better than ensembles (unlike in the one-layer case). 

\subsection{Supplementary to the Rocket Booster Simulation}  \label{appendix:nasadata}
\begin{figure}[ht!]
    \centering
    \includegraphics[clip, trim=0.0cm 3cm 0.0cm 0.0cm,width=0.8\linewidth]{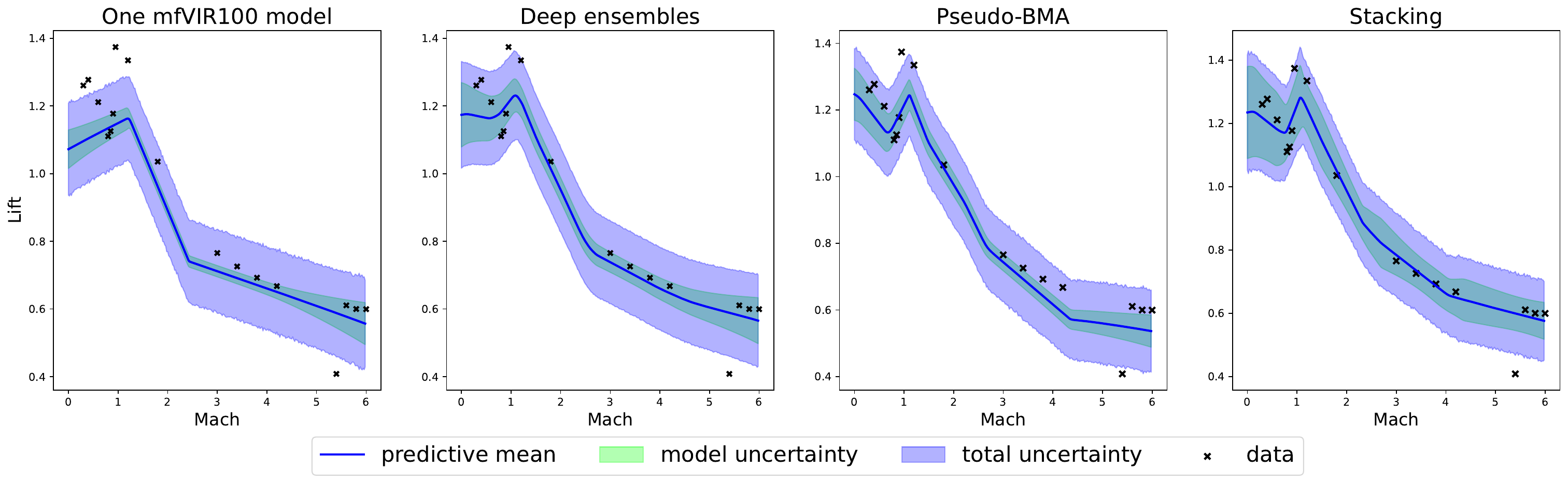}
      \includegraphics[clip, trim=0.0cm 3cm 0.0cm 0.0cm,width=0.8\linewidth]{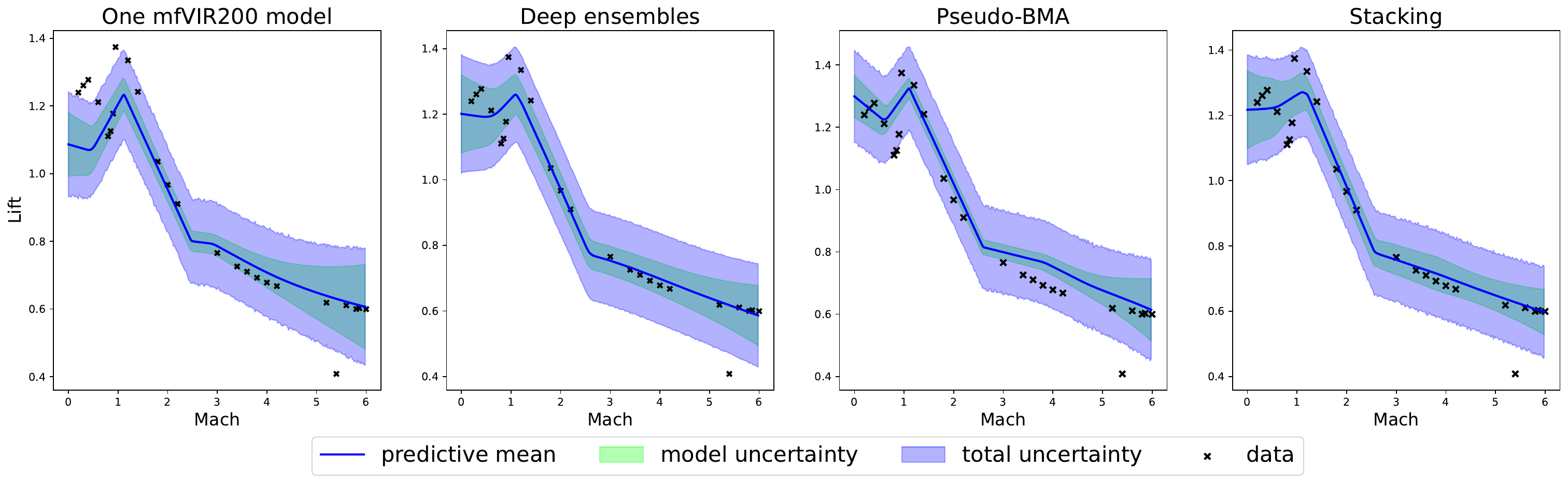}
    \includegraphics[clip, trim=0.0cm 3cm 0.0cm 0.0cm,width=0.8\linewidth]{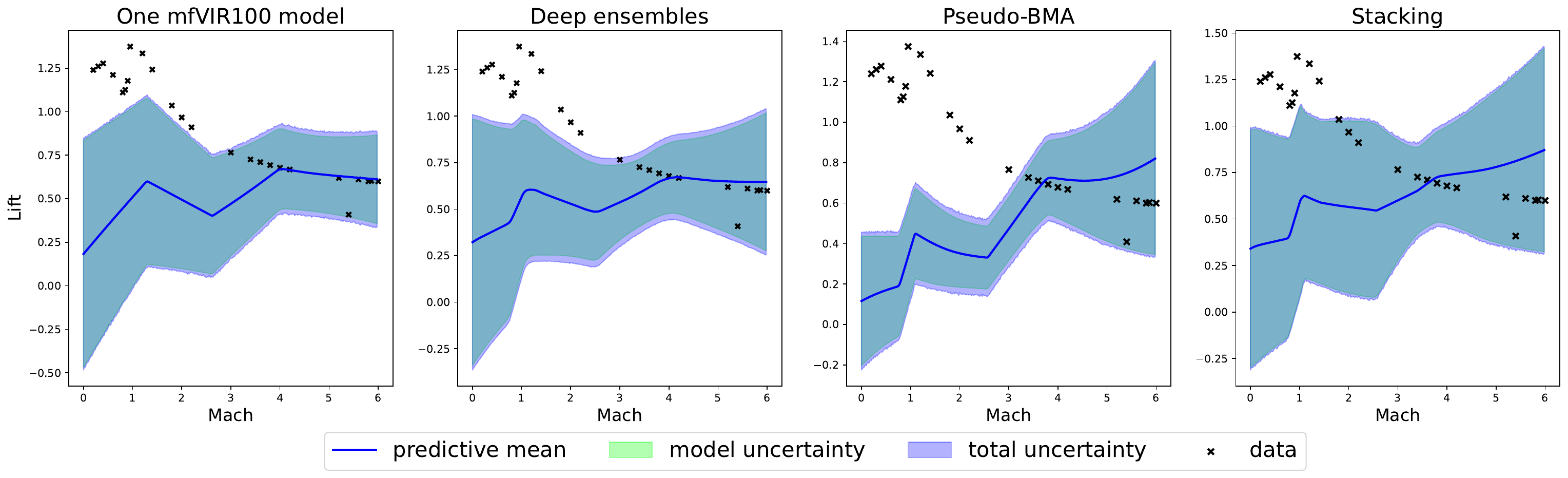}
        \includegraphics[width=0.8\linewidth]{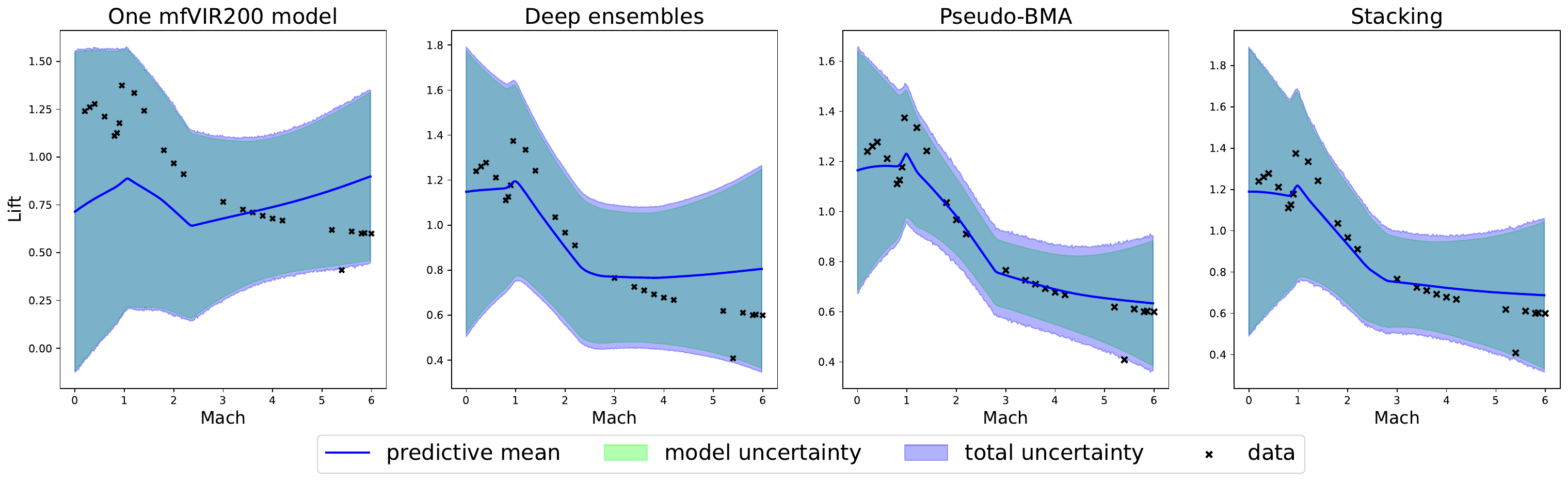}
    \caption{Slices of predicted lift obtained by a single model and by ensembling, stacking and pseudo-BMA when applied to mfVIR100 and mfVIR200 with Gaussian priors in the non-OOD (top figures)  and 'complement-distributions'  OOD (bottom figures) tasks. Stacking outperforms other methodologies.}
    \label{fig:nasa_slicesup}
\end{figure}

When considering the Langley Glide-Back Booster simulator data \citep{rogers2003automated}, we evaluated the performance of the mfVIR for various widths and in both non-OOD and 'complement-distributions' tasks. Here we provide additional figures for models with 100 and 200 hidden units illustrating the slices of predicting lift, with Mach on the x-axis and angles $\alpha$ and $\beta$ fixed to 25 and 4, respectively.

\end{document}